%% file: neurips_2025.tex
\documentclass{article}

\PassOptionsToPackage{numbers, compress}{natbib}
\usepackage[preprint]{neurips_2025}

\input{macros}
\title{LoMix: Learnable Weighted Multi‑Scale Logits Mixing for Medical Image Segmentation}

\author{%
Md Mostafijur Rahman \quad Radu Marculescu \\
Department of Electrical and Computer Engineering\\The University of Texas at Austin\\Austin, TX 78703 \\
\texttt{\{mostafijur.rahman,radum\}@utexas.edu}
}

\begin{document}

\maketitle

%%%%%%%%%%%%%%%%%%%%%%%%%%%%%%%%%%%%%%%%%%%%%%%%%%%%%%%%%%%%%%%%%%%%%%%%%%%%%%%%%%%%%%%%%%%%%%%%%%%%%
\input{sections/0.abstracts}
\input{sections/1.introduction}
\input{sections/2.related_works}
\input{sections/3.technique}

\input{sections/4.experiments}

\input{sections/5.ablations}
\input{sections/6.conclusion}
%%%%%%%%%%%%%%%%%%%%%%%%%%%%%%%%%%%%%%%%%%%%%%%%%%%%%%%%%%%%%%%%%%%%%%%%%%%%%%%%%%%%%%%%%%%%%%%%%%%%%

%\bibliographystyle{splncs04}
%\bibliography{main}

%%%%%%%%%%%%%%%%%%%%%%%%%%%%%%%%%%%%%%%%%%%%%%%%%%%%%%%%%%%%%%%%%%%%%%%%%%%%%%%%%%%%%%%%%%%%%%%%%%%%%

\newpage

%\clearpage
\setcounter{page}{1}
% ---------- 1.  Number figures as S.1, S.2, … ----------
\makeatletter
\setcounter{figure}{0}
\renewcommand{\thefigure}{S.\arabic{figure}}

% ---------- 2.  Number tables as S1, S2, … -------------
\setcounter{table}{0}
\renewcommand{\thetable}{S\arabic{table}}
\makeatother

\appendix

\section{Appendix / Supplementary Material}

\subsection{Training Algorithm} Algorithm~\ref{alg:training} outlines the training procedure for our U-shaped segmentation network with the LoMix module. We train end-to-end with AdamW \cite{loshchilov2017decoupled}, updating both the usual network parameters and the loss weight parameters $\alpha$. Pseudocode is given in Algorithm \ref{alg:training} below.

\begin{algorithm}[b]
\caption{Training with Logits Mixing (LoMix)}
\label{alg:training}
\begin{algorithmic}[1]
  \Input{%
    Labeled image dataset $\mathcal{D}=\{(I_n,Y_n)\}_{n=1}^{N}$; number of decoder stages $L$; network parameters $\Theta\!$ (encoder, decoders); fusion–weight scalars $\{\alpha_u\}\leftarrow 0$ for every original or mutant logit $P_u$
  }
  \Output{Optimized model $\Theta$ and fusion-weight scalars\ $\{\alpha_u\}$}
  \State Initialize optimizer $\mathcal{O}$ for $(\Theta,\{\alpha_u\})$
  \While{training not converged}
     \State Sample mini-batch $\{(I_b,Y_b)\}_{b=1}^{B}\!\subset\!\mathcal{D}$
     \State $\{P_\ell\}_{\ell=1}^{L}\leftarrow f_{\Theta}(I_b)$ \Comment{upsampled stage logits}
     \State $\mathcal{P}_{\mathrm{orig}}\leftarrow\{P_1,\dots,P_L\}$,\;$\mathcal{P}_{\mathrm{mut}}\leftarrow\emptyset$
     \ForAll{non-empty subsets $S\subseteq\{1,\dots,L\}$ with $|S|\ge2$}
         \State Generate fused logits $P_S^{(add)}$, $P_S^{(mult)}$, $P_S^{(\mathrm{cat})}$, $P_S^{(\mathrm{awf})}$ via
               Eqs.~\ref{eq:addition_fusion}–\ref{eq:awf_fusion}
         \State $\mathcal{P}_{\mathrm{mut}}\leftarrow\mathcal{P}_{\mathrm{mut}}\cup\!\bigl\{P_S^{(add)} ,P_S^{(mult)},P_S^{(\mathrm{cat})},P_S^{(\mathrm{awf})}\bigr\}$
     \EndFor
     \State $\mathcal{L}_{total}\leftarrow 0$
     \ForAll{$P_u\in\mathcal{P}_{\mathrm{orig}}\cup\mathcal{P}_{\mathrm{mut}}$}
         \State $w_u\leftarrow \mathrm{softplus}(\alpha_u)$  \Comment{NAS-inspired, unconstrained}
         \State $\mathcal{L}_{total}\; \leftarrow\; \mathcal{L}_{total} + w_u \,\mathcal{L}_{\text{seg}}\!\bigl(P_u,Y_b\bigr)$
     \EndFor
     \State Update $(\Theta,\{\alpha_u\}) \leftarrow \mathcal{O}\!\bigl(\nabla_{\Theta,\alpha}\mathcal{L}_{total}\bigr)$
  \EndWhile
  \State\Return $\Theta,\{\alpha_u\}$
\end{algorithmic}
\end{algorithm}

During training, the network learns to minimize the total loss by improving all logits. As training progresses, the $\mathrm{softplus}(\alpha)$ weights will adjust – for example, if the attention-fused output consistently yields lower error than others, its corresponding $\alpha_{ij}^{\text{awf}}$ may increase, giving it more influence.

In the end, we have multiple trained decoder branches and fusion modules. For inference, one can either use output of the single best-performing decoder stage or combine the decoder outputs via one of the fusion strategies (e.g., attention fusion or a simple average) to produce the final segmentation. 

Thanks to our LoMix supervised training, all these outputs are optimized to be accurate and complementary. The result is a segmentation model that embodies an ensemble of experts, trained jointly in a principled manner to maximize the segmentation quality.

\subsection{Advantages of LoMix} Our LoMix-enhanced segmentation has several notable advantages:

\begin{itemize}
    
\item \textbf{Interpretability:} The learned weights $\mathrm{softplus}(\alpha_i)$ and $\mathrm{softplus}(\alpha_S^{op})$ provide insights into the importance of each decoder output and fusion type. For instance, if the addition-fusion outputs receive a high weight, it means that the simple aggregated prediction is consistently effective; if a particular logits’s weight drops close to zero, then the system concludes that that particular logit is not helpful. Examining these weights can thus reveal which combinations of predictions the model finds most useful, thus offering a peek into the ensemble strategy learned by the network.

\item \textbf{Adaptability:} Because the loss weights are learned, the framework adapts to different datasets and scenarios. The model can allocate more weight to certain predictions if the data benefit from that fusion. For example, on data where one decoder stage consistently outperforms the others, the learning process can put more emphasis to that decoder stage (and possibly its fused outputs) to optimize performance. Conversely, if all decoders are needed (say each captures a different class or scale), then the weights can remain distributed. This adaptability alleviates the need for manual hyperparameter tuning of multi-output losses for each new application.

\item \textbf{Training Stability and Regularization:} Supervising multiple predictions (original and fused) acts as an implicit regularizer and stabilizer. Indeed, it is less likely that the network will overfit or get stuck in a poor local minimum because each decoder stage is guided by its own loss and by the fused losses that tie all decoder stages together. If one decoder begins to make mistakes, the others (and their combinations) still provide correct feedback, preventing the entire model from drifting off. Additionally, the learned weighting further stabilizes training by reducing the impact of any particularly noisy loss term: if a fused output is extremely erroneous at the start, its weight can adjust downward, preventing it from exploding the gradient. Overall, we observed that LoMix yields faster convergence and more robust training than a single-output or manually deep-supervised counterpart, thanks to these effects.

\end{itemize}

\subsection{Why Softplus in NAS-inspired Weight Learning?}
We choose Softplus (see Eq. \ref{eq:softplus_weights}) rather than alternatives like Softmax or explicit normalization for several reasons:

\begin{itemize}
    
\item \textbf{No Sum Constraint.} Softmax would constrain the weights to sum to one, coupling them and forcing a distribution over outputs. This would prevent the model from independently suppressing a noisy output (as reducing one weight necessitates increasing others). In contrast, Softplus outputs are independent and unbounded, so each weight can shrink toward zero (effectively ignoring that logit) or grow arbitrarily without affecting the sum of other weights.
\item \textbf{Strict Positivity.} Softplus guarantees $\alpha>0$ smoothly, unlike ReLU which could produce exact zeros or sigmoid which would bound weights in $(0,1)$. Positive weights ensure each loss term contributes non-negatively.
\item \textbf{Smooth Gradients.} Softplus is smooth and has non-vanishing derivatives for all inputs, which stabilizes learning of the weight parameters. A hard normalization or clipping could yield zero gradients for some ranges.

\end{itemize}

\subsection{Datasets}
We evaluate the LoMix's efficacy across seven datasets covering six segmentation tasks. Our two multi-class segmentation datasets are Synapse Multi-organs \footnote{\href{https://www.synapse.org/\#!Synapse:syn3193805/wiki/217789}{https://www.synapse.org/\#!Synapse:syn3193805/wiki/217789 }} and ACDC cardiac organs \footnote{\href{https://www.creatis.insa-lyon.fr/Challenge/acdc/}{https://www.creatis.insa-lyon.fr/Challenge/acdc/}}. The Synapse multi-organ dataset is used for abdominal organ segmentation and includes 30 abdominal CT scans with 3,779 axial slices of $512 \times 512$ pixels. Following the TransUNet \cite{chen2021transunet}, 18 scans (2,212 slices) are used for training and 12 for validation/testing. We segment eight organs: aorta, gallbladder, left kidney, right kidney, liver, pancreas, spleen, and stomach. For cardiac organ segmentation, the ACDC dataset contains 100 cardiac MRI scans segmented into three sub-organs: right ventricle (RV), myocardium (MYO), and left ventricle (LV). We follow the TransUNet protocol using 70 cases (1,930 slices) for training, 10 for validation, and 20 for testing. Our binary breast cancer segmentation dataset, BUSI \cite{al2020dataset} contains 647 images: 437 benign and 210 malignant. Our skin lesion segmentation dataset. Our three polyp segmentation datasets are Kvasir \cite{jha2019kvasir} (1,000 images), ClinicDB [3] (612 images), CVC-ColonDB \cite{tajbakhsh2015automated} (379 images), and ETIS-LaribPolypDB \cite{silva2014toward} (196 images). Furthermore, we use ISIC2018 \cite{codella2019skin} (2,594 images) for skin lesion segmentation. We use 80\% of the data for training, 10\% for validation, and 10\% for testing in BUSI, Kvasir, CVC-ColonDB, ETIS-LaribPolypDB, and ISIC2018 datasets. 

\subsection{Dataset Specific Implementation Details}
For multi-class segmentation in Synapse Multi-organs and ACDC datasets, we use an input size of $224\times224$, and optimize the combined Cross-entropy ($\beta$=0.3) + DICE ($\gamma$=0.7) loss. We train models for 300 and 400 epochs with a batch size of 6 and 12 for Synapse and ACDC datasets, respectively. The image dimensions are set to $256\times256$ pixels for the BUSI and ISIC2018 datasets, while the image dimensions are set to $352\times352$ pixels for the polyp datasets (Kvasir, CVC-ColonDB, ETIS-LaribPolypDB), respectively. We utilize a multi-scale training approach, with scales of \{0.75, 1.0, 1.25\} and no augmentation. We use a hybrid weighted BinaryCrossEntropy (BCE) with a weighted Intersection over Union (IoU) loss (1:1) and train models for 200 epochs with batches of 16 for BUSI, ISIC2018, Kvasir, CVC-ColonDB, and ETIS-LaribPolypDB. We employ random rotation and flipping as data augmentation methods in all of our experiments except the BUSI dataset. We save the best model based on validation DICE score in all datasets and report DICE score on testsets except the Synapse Multi-organ dataset. Only the last stage prediction is chosen as final segmentation output for Synapse Multi-organ and ACDC datasets, while the predictions from all four stages are summed together to produce the final segmentation map in the BUSI, ISIC 2018, Kvasir, CVC-ColonDB, and ETIS-LaribPolypDB datasets. 

\begin{figure}[t]
  \centering
  \begin{subfigure}[t]{0.49\linewidth}
    \centering
    \includegraphics[width=\linewidth]{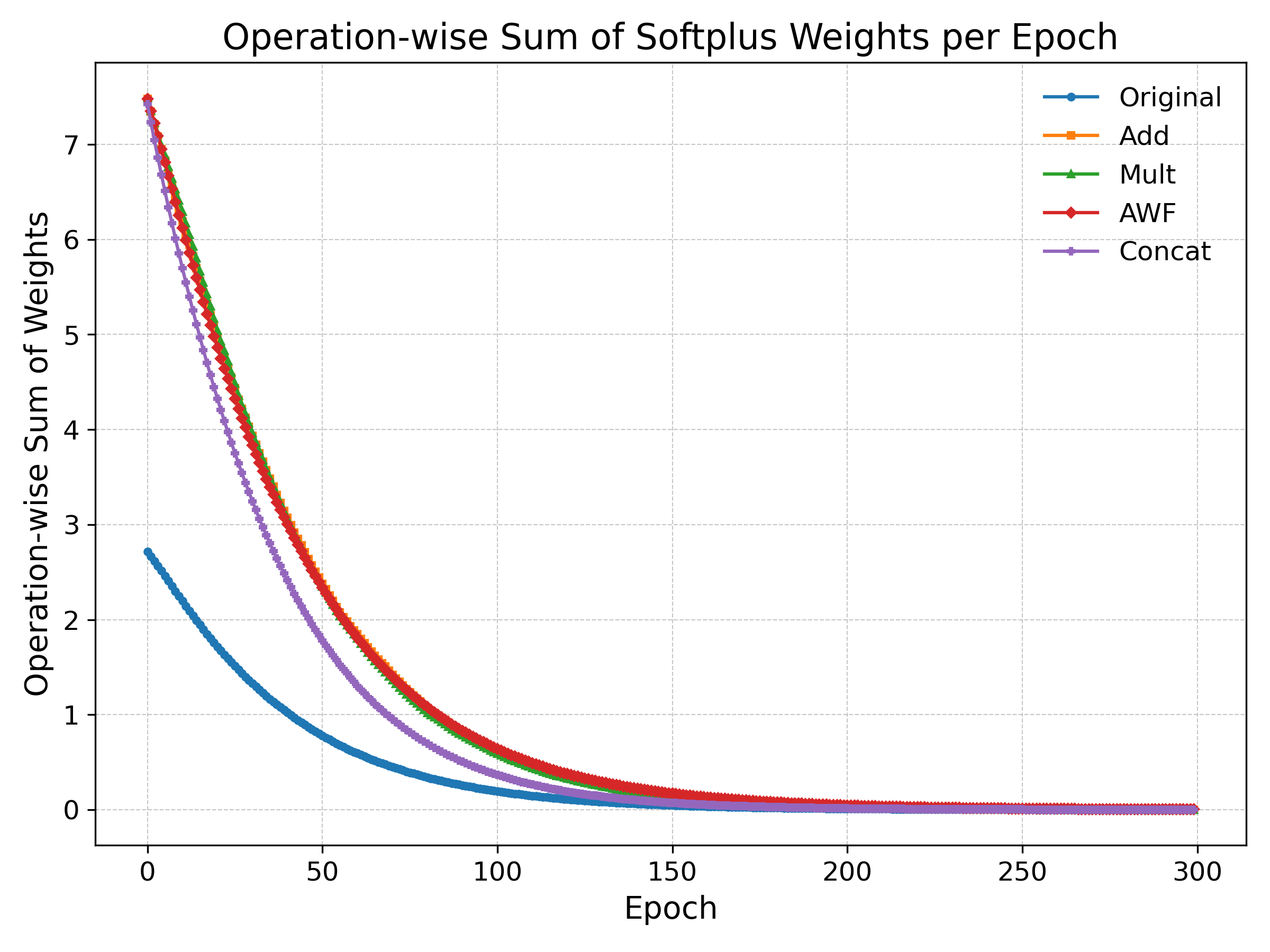}
   % \vspace{-0.4cm}
    \caption{Linear scale}
  \end{subfigure}\hfill
  \begin{subfigure}[t]{0.49\linewidth}
    \centering
    \includegraphics[width=\linewidth]{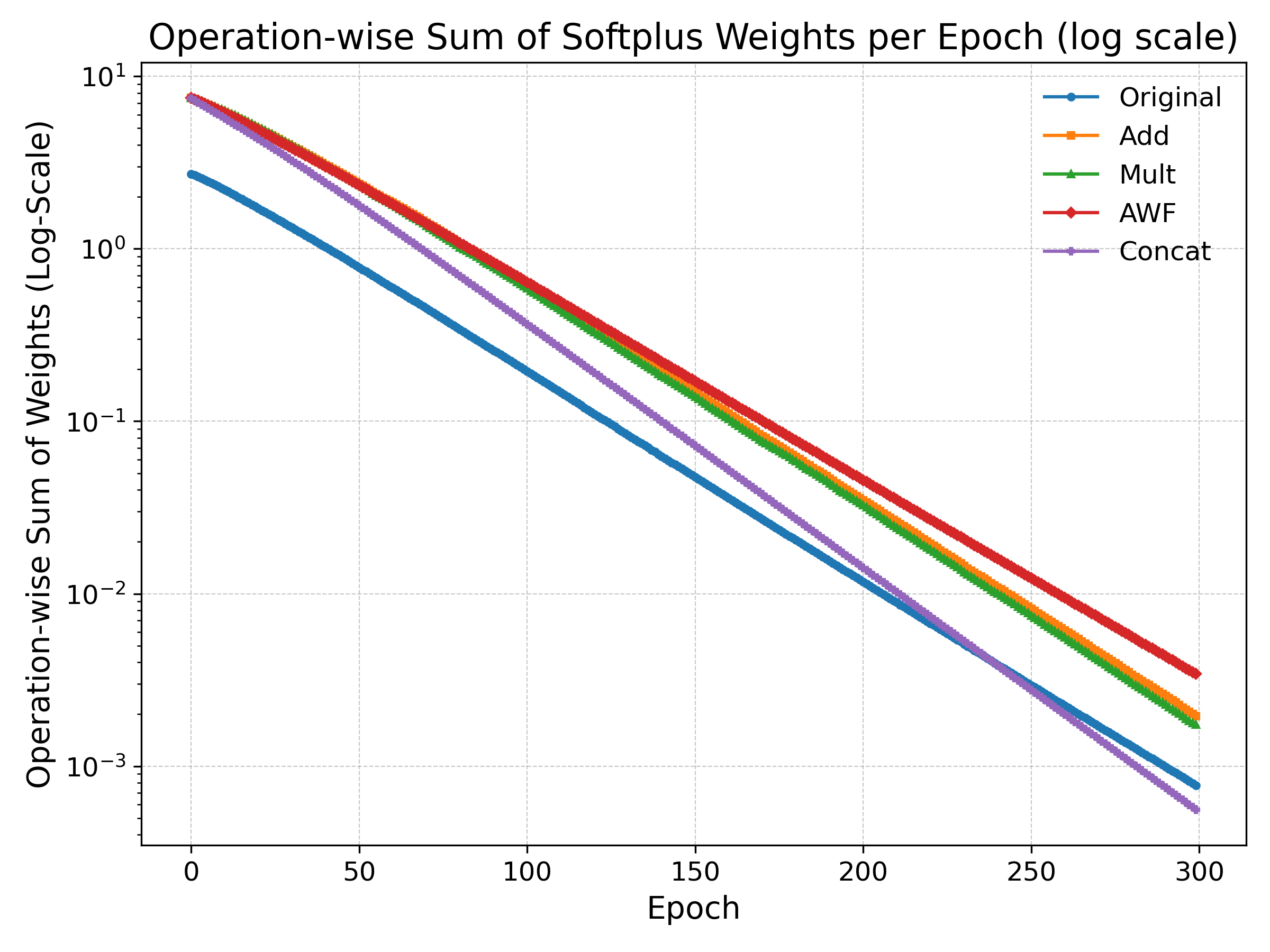}
   % \vspace{-0.4cm}
    \caption{Log$_{10}$ scale}
  \end{subfigure}
  %
  %\vspace{-0.2cm}
  \caption{Operation-wise sum of softplus loss weights during training. Each curve aggregates all logits produced by the same fusion family (Original, Add, Mul, WF, Concat) in the PVT-EMCAD-B2 + LoMix run. (a) softplus weight values in linear scale, (b) the same values in log-scale.}
  \label{fig:group_weight_curves}
  \vspace{-0.3cm}
\end{figure}

\begin{figure}%[t]
    \centering
    %---------------------------------------------------------
    \subfloat[Linear scale\label{fig:softplus_linear}]{
        \includegraphics[width=0.48\textwidth]{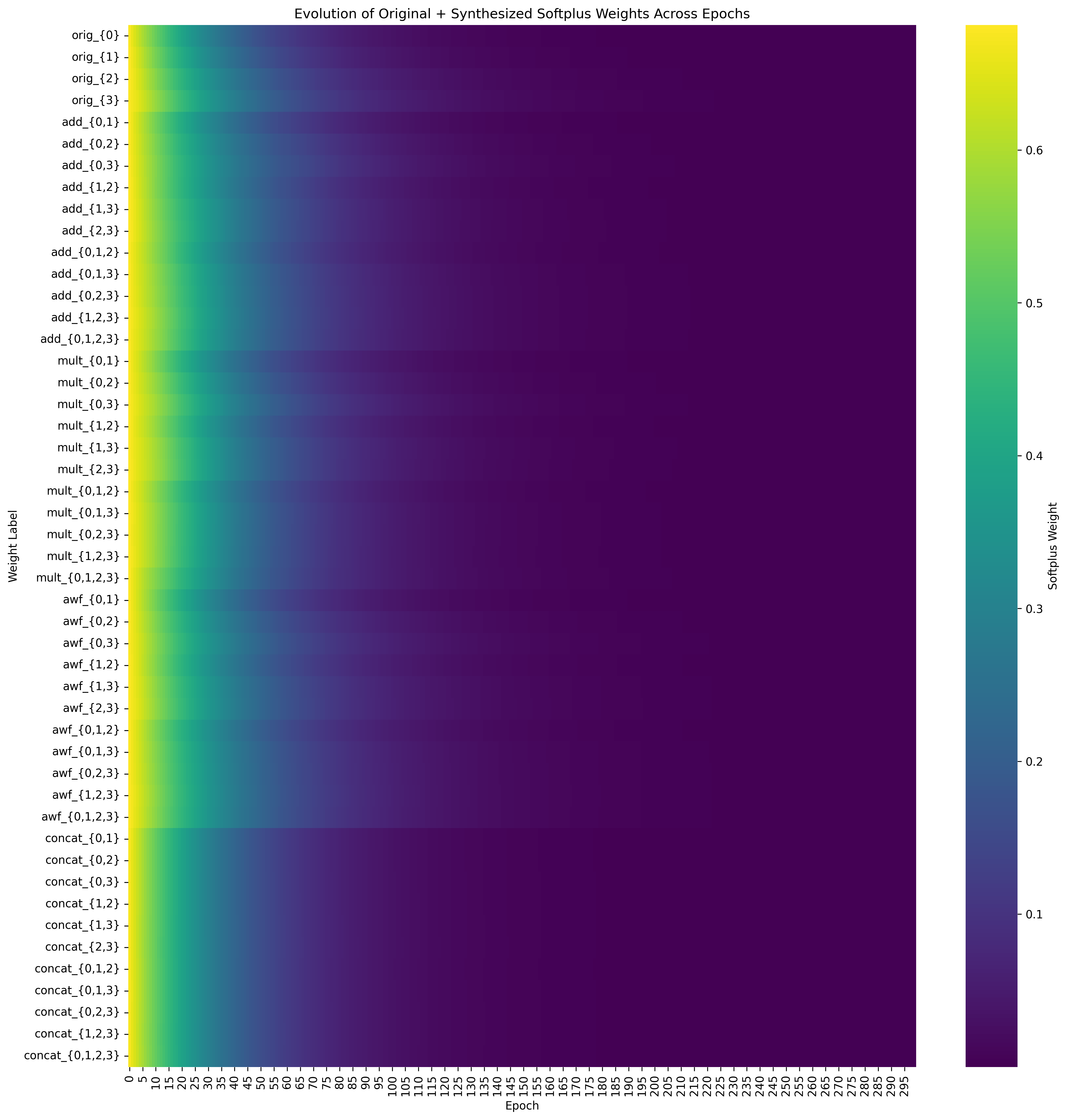}}
    \hfill
    \subfloat[Log$_{10}$ scale\label{fig:softplus_log}]{
        \includegraphics[width=0.505\textwidth]{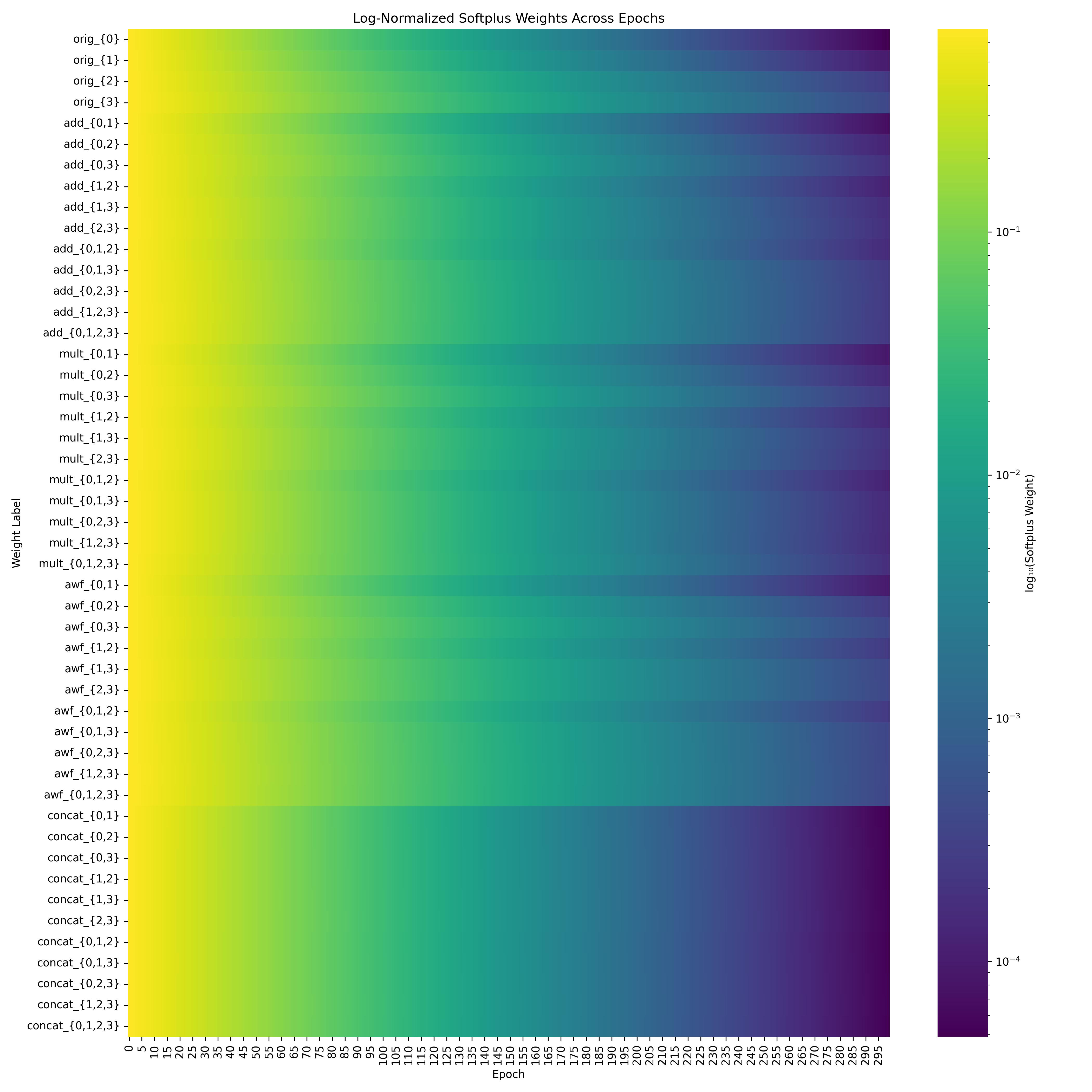}}
    %---------------------------------------------------------
    %\vspace{-0.25cm}
    \caption{Evolution of softplus weights over training epochs in Synapse 8-organ segmentation. (a) softplus values in linear-scale, (b) the same data with a logarithmic color normalization, revealing the relative ordering of very small weights and confirming that a few predictions dominate the loss while many others are softly suppressed.}
    \label{fig:softplus_heatmaps}
    %\vspace{-0.6cm}
\end{figure}

\subsection{Weight Dynamics and Interpretability}
\label{ssec:weight_dynamics}

Figure \ref{fig:group_weight_curves} shows that LoMix reallocates supervision away from the four original logits toward the far larger pool of fused logits. By epoch 50, the original logits hold $<10\%$ of the total weight, while the learnable fusions, especially attention-weighted fusion (AWF) and multiplication (Mult) remain dominant for the next 250 epochs. Addition and concatenation also preserve non-trivial weight, illustrating that every operator family contributes useful gradients. The log-scale panel highlights near-perfect exponential decay for all groups, with parallel slopes indicating that LoMix balances them proportionally rather than suppressing any single operator outright. Together, these curves confirm that the NAS-inspired optimization discovers a nuanced, multi-operator loss distribution in which learnable fusions drive most of the training signal, yet fixed arithmetic fusions and even the original heads are still retained to provide complementary guidance.

Figure \ref{fig:softplus_heatmaps} visualizes the softplus loss weights that LoMix learns for \emph{all} 48 supervised logits (4 originals + 44 fusions) over 300 epochs. In the linear-scale map (Figure \ref{fig:softplus_heatmaps}a), the vast majority of weights collapse to closer to 0 after 50 epochs, while a small cluster—mainly the finest (last layer) original logit and a few mutants created by \textbf{AWF}, \textbf{Add}, \textbf{Mul}, and \textbf{Concat}—retain higher weights. The log-scale view (Figure \ref{fig:softplus_heatmaps}b) makes the hierarchy clearer: the finest grain and attention-weighted fusion (awf) maps dominates over other original or synthetic maps. Thus, the differentiable search automatically reduces 48 supervision maps down to a compact, high-impact subset while discarding redundant logits.

\begin{figure}%[t]
  \centering
  \begin{subfigure}[t]{0.49\linewidth}
    \centering
    \includegraphics[width=\linewidth]{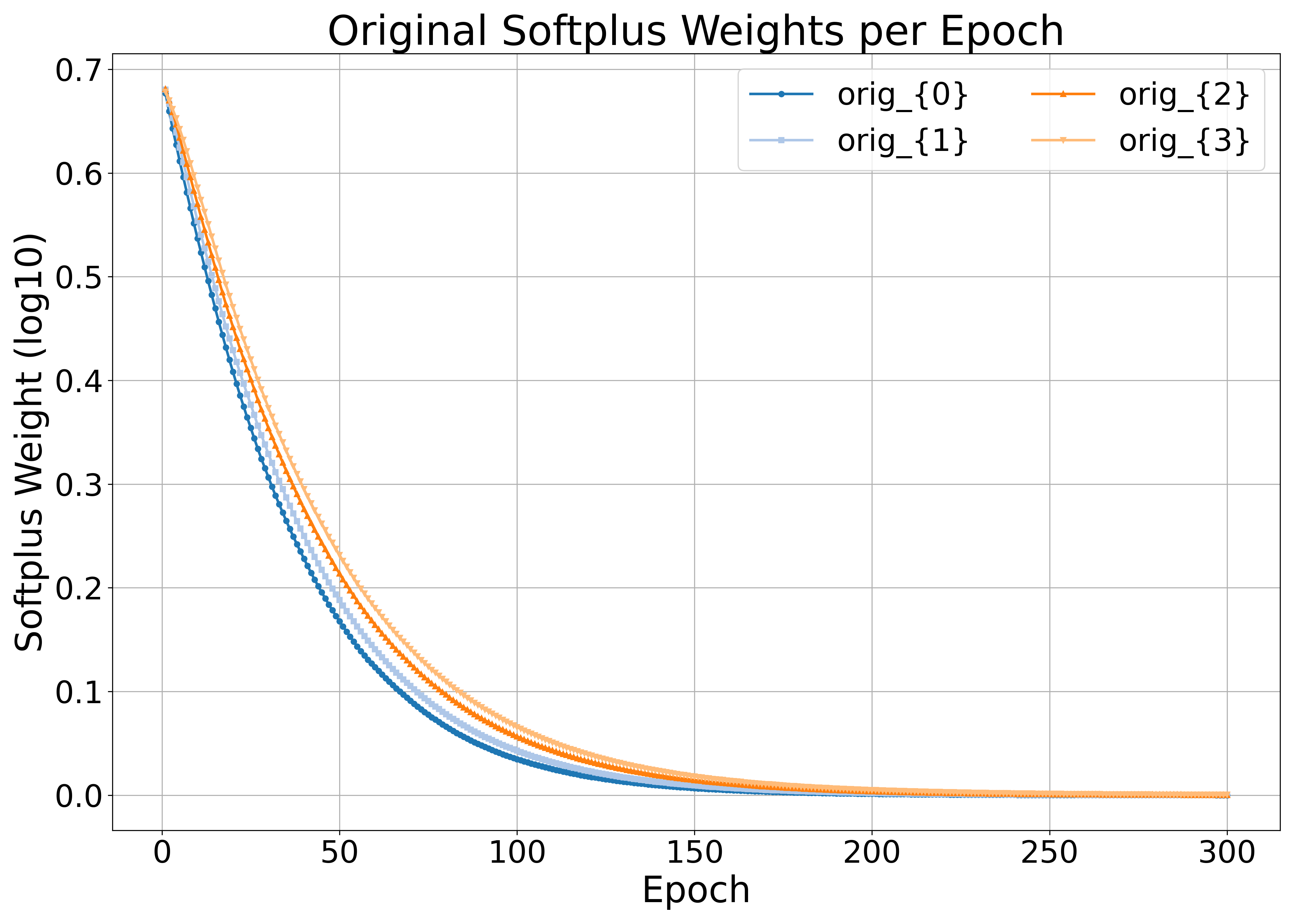}
    \caption{Linear scale}
  \end{subfigure}\hfill
  \begin{subfigure}[t]{0.49\linewidth}
    \centering
    \includegraphics[width=\linewidth]{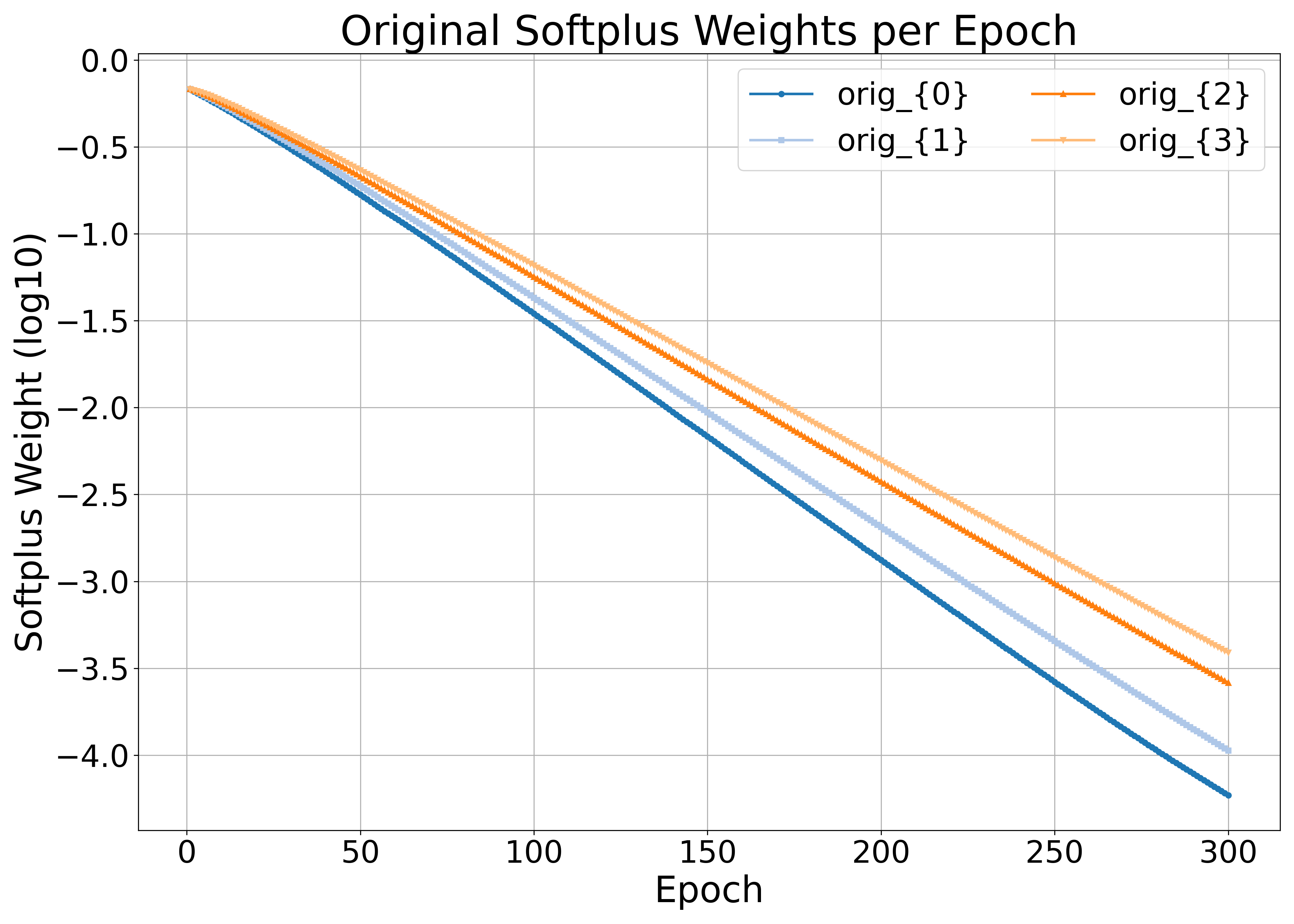}
    \caption{Log$_{10}$ scale}
  \end{subfigure}
  %
  %\vspace{-0.25cm}
  \caption{Evolution of softplus loss weights for the four original logits. Both panels track the same data over 300 training epochs; (a) the left panel plot the softplus values in linear scale; (b) the right plot rescales the \emph{y}-axis logarithmically to expose tiny values.}
  \label{fig:orig_weight_curves}
  %\vspace{-0.3cm}
\end{figure}

\begin{figure}%[t]
  \centering
  \begin{subfigure}[t]{0.49\linewidth}
    \centering
    \includegraphics[width=\linewidth]{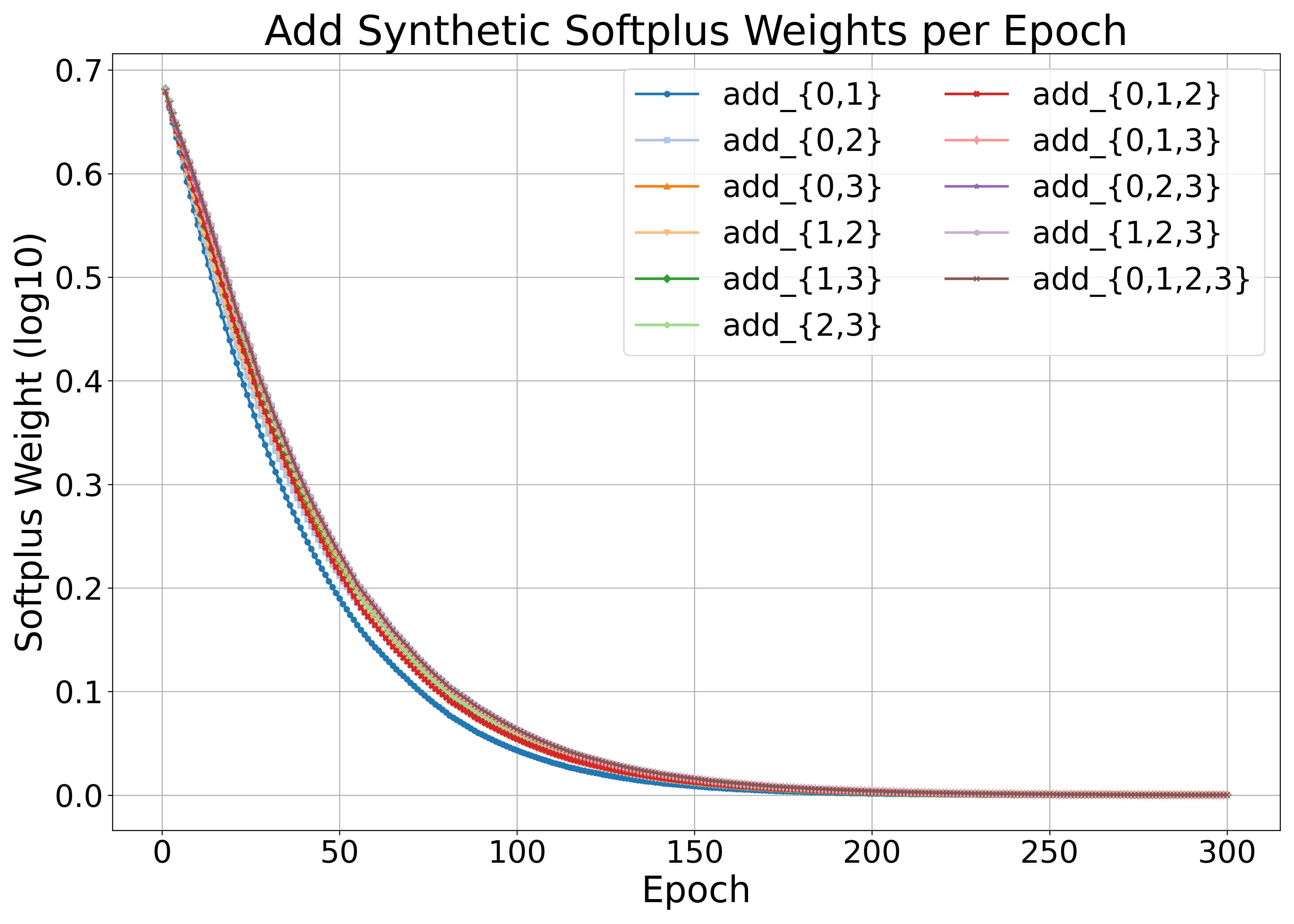}
    \caption{Linear scale}
  \end{subfigure}\hfill
  \begin{subfigure}[t]{0.49\linewidth}
    \centering
    \includegraphics[width=\linewidth]{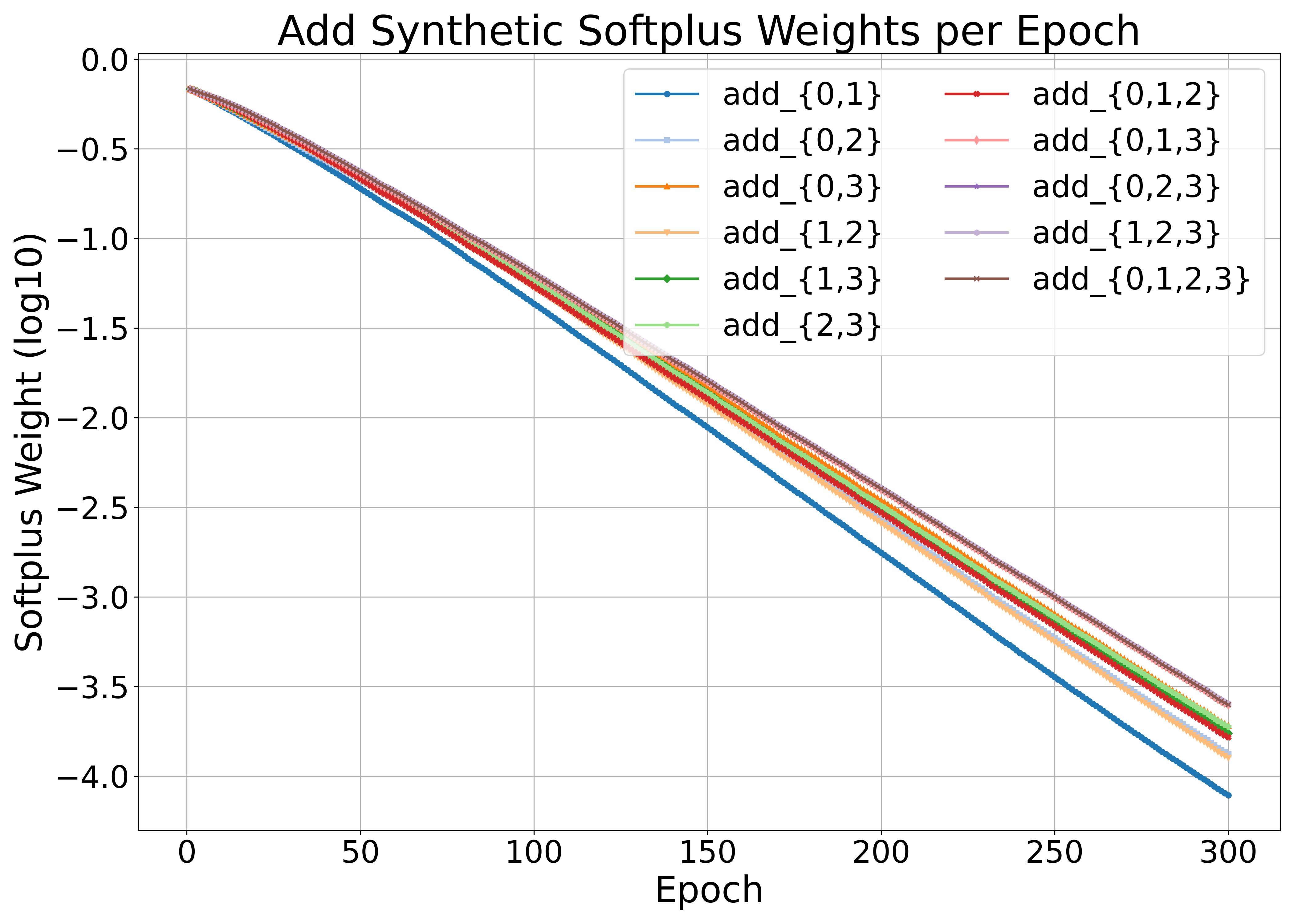}
    \caption{Log$_{10}$ scale}
  \end{subfigure}
  %
  %\vspace{-0.2cm}
  \caption{Per-subset softplus weights for the \emph{addition} operation during training. Curves correspond to every additive mixture of decoder logits. Both the (a) linear- and (b) log–scaled plots reveal how the NAS-inspired optimization reallocates loss weight across epochs.}
  \label{fig:add_weight_curves}
  %\vspace{-0.4cm}
\end{figure}

\begin{figure}%[t]
%\vspace{-0.3cm}
  \centering
  \begin{subfigure}[t]{0.49\linewidth}
    \centering
    \includegraphics[width=\linewidth]{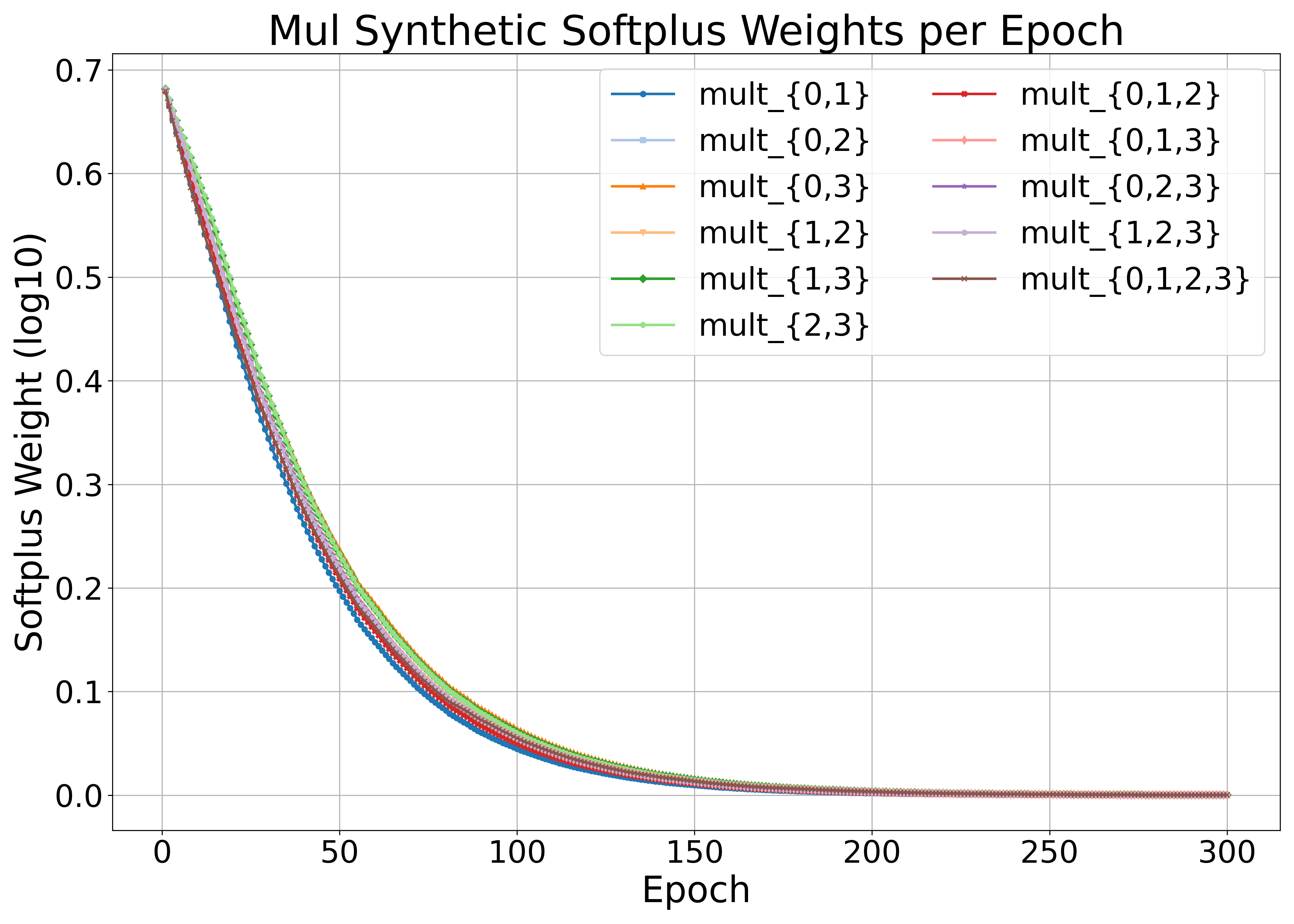}
    \caption{Linear scale}
  \end{subfigure}\hfill
  \begin{subfigure}[t]{0.49\linewidth}
    \centering
    \includegraphics[width=\linewidth]{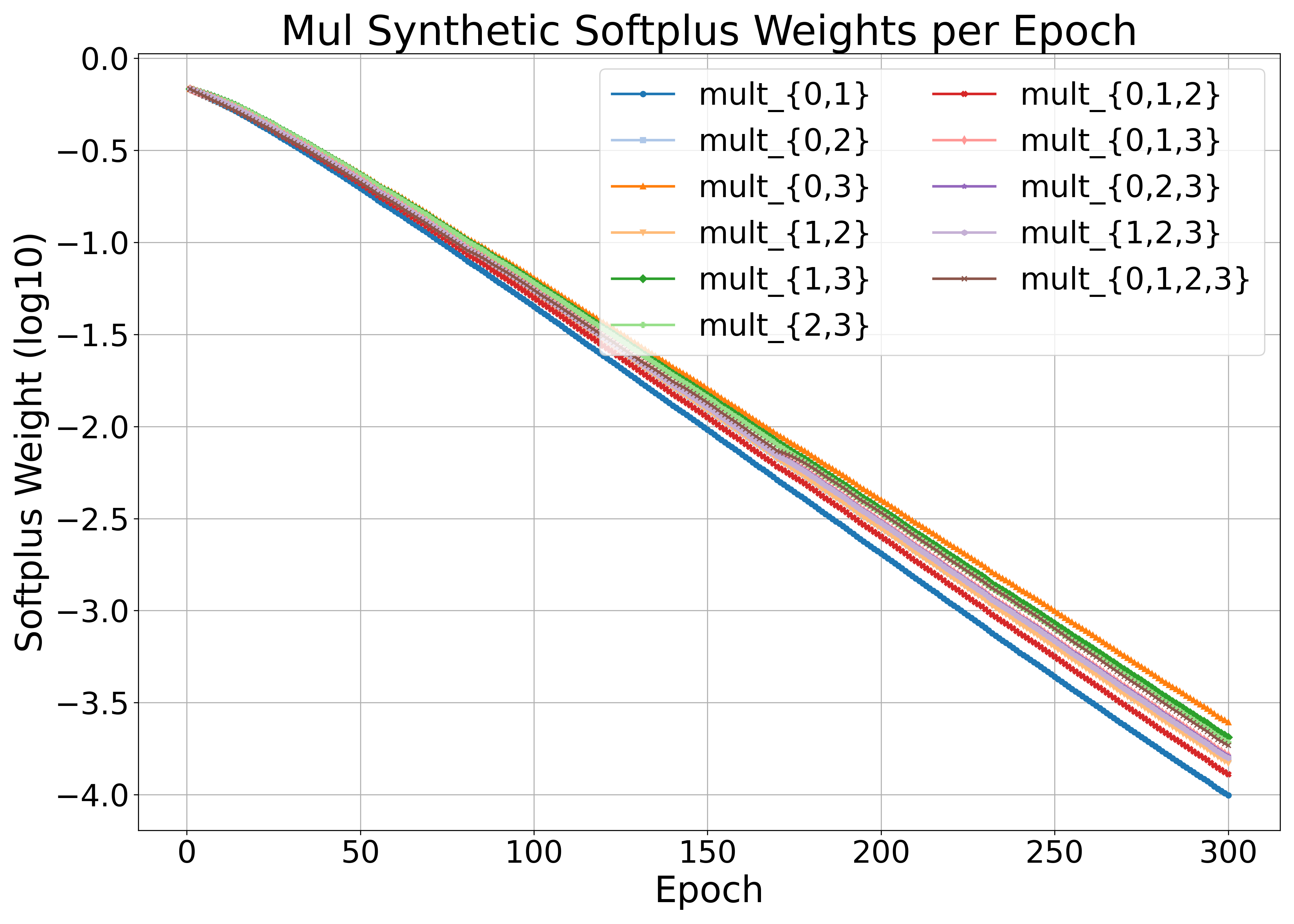}
    \caption{Log$_{10}$ scale}
  \end{subfigure}
  %
  %\vspace{-0.2cm}
  \caption{Per-subset softplus weights for the \emph{multiplication} operation during training. Curves correspond to every multiplication mixture of decoder logits. Both the (a) linear- and (b) log–scaled plots reveal how the NAS-inspired optimization reallocates loss weight across epochs.}
  \label{fig:mul_weight_curves}
  %\vspace{-0.7cm}
\end{figure}

\textbf{Original.} Figure \ref{fig:orig_weight_curves} shows how LoMix optimizes the weights to four original decoder outputs during training. Both linear-scale (Figure \ref{fig:orig_weight_curves}a)and log-scale (Figure \ref{fig:orig_weight_curves}b) reveal that the fine-scale heads ($orig\_{2}$ / $orig_{3}$) retaining more influence than the coarser ones during all training epochs. This trend mirrors the full heatmap in Figure \ref{fig:softplus_heatmaps}: the NAS-inspired optimization quickly reallocates the supervision to a handful of high-utility fused logits, allowing LoMix to concentrate the gradient signal where it is most beneficial.

\begin{figure}%[t]
%\vspace{-0.3cm}
  \centering
  \begin{subfigure}[t]{0.49\linewidth}
    \centering
    \includegraphics[width=\linewidth]{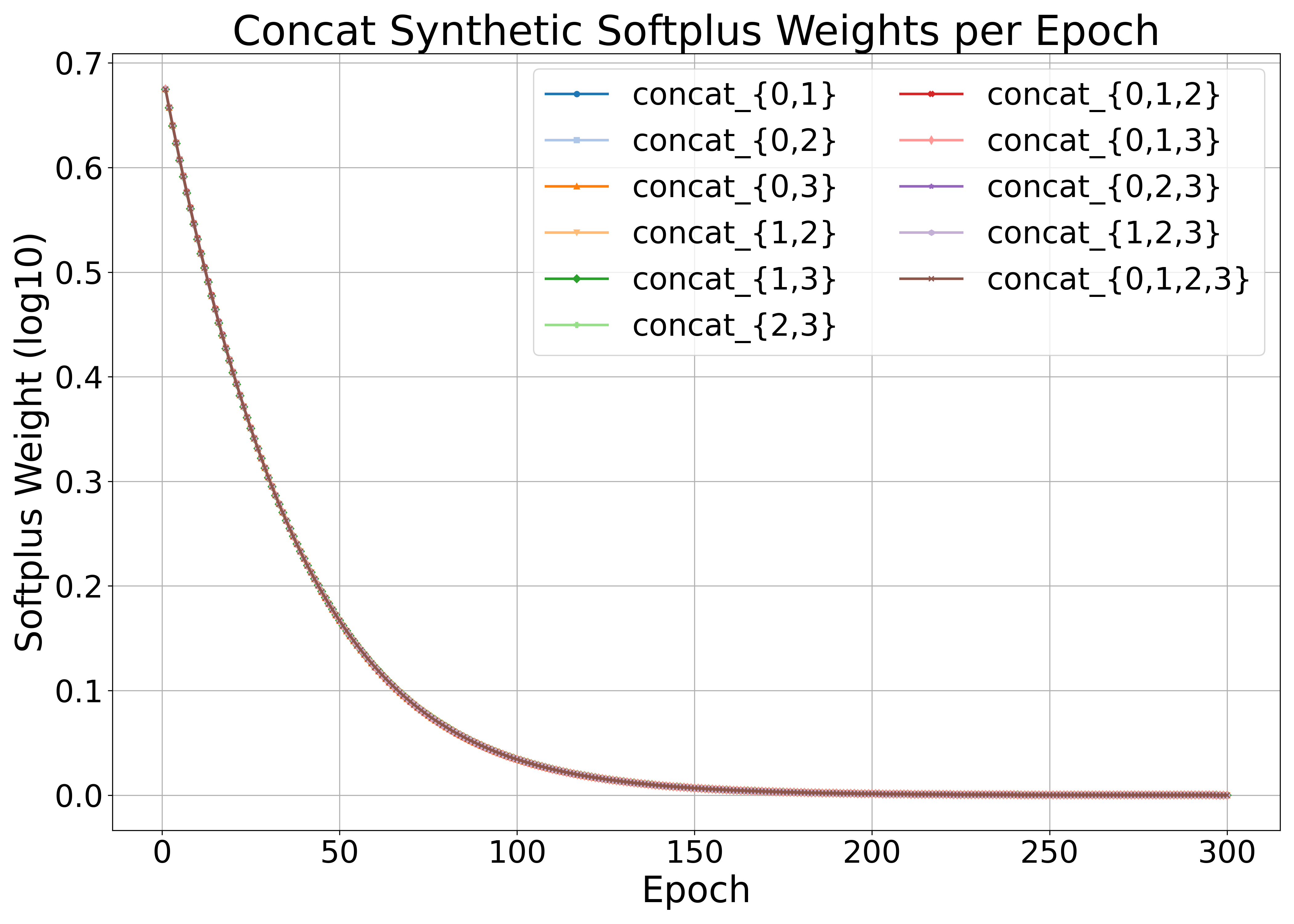}
    \caption{Linear scale}
  \end{subfigure}\hfill
  \begin{subfigure}[t]{0.49\linewidth}
    \centering
    \includegraphics[width=\linewidth]{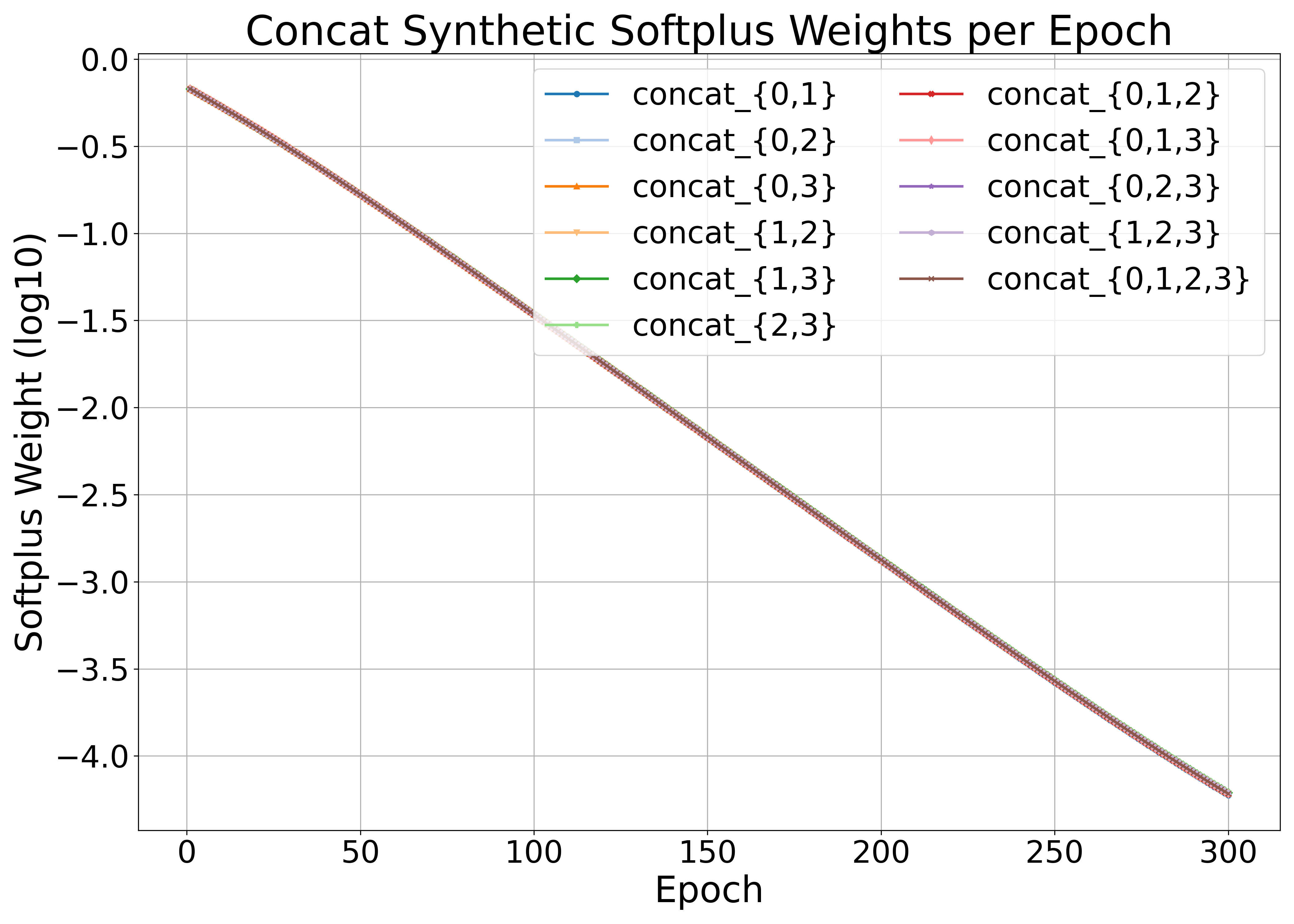}
    \caption{Log$_{10}$ scale}
  \end{subfigure}
  %
  %\vspace{-0.2cm}
  \caption{Per-subset softplus weights for the \emph{concatenation} operation during training. Curves correspond to every concatenation mixture of decoder logits. Both the (a) linear- and (b) log–scaled plots reveal how the NAS-inspired optimization reallocates loss weight across epochs.}
  \label{fig:concat_weight_curves}
  %\vspace{-0.7cm}
\end{figure}

\begin{figure}%[t]
%\vspace{-0.3cm}
  \centering
  \begin{subfigure}[t]{0.49\linewidth}
    \centering
    \includegraphics[width=\linewidth]{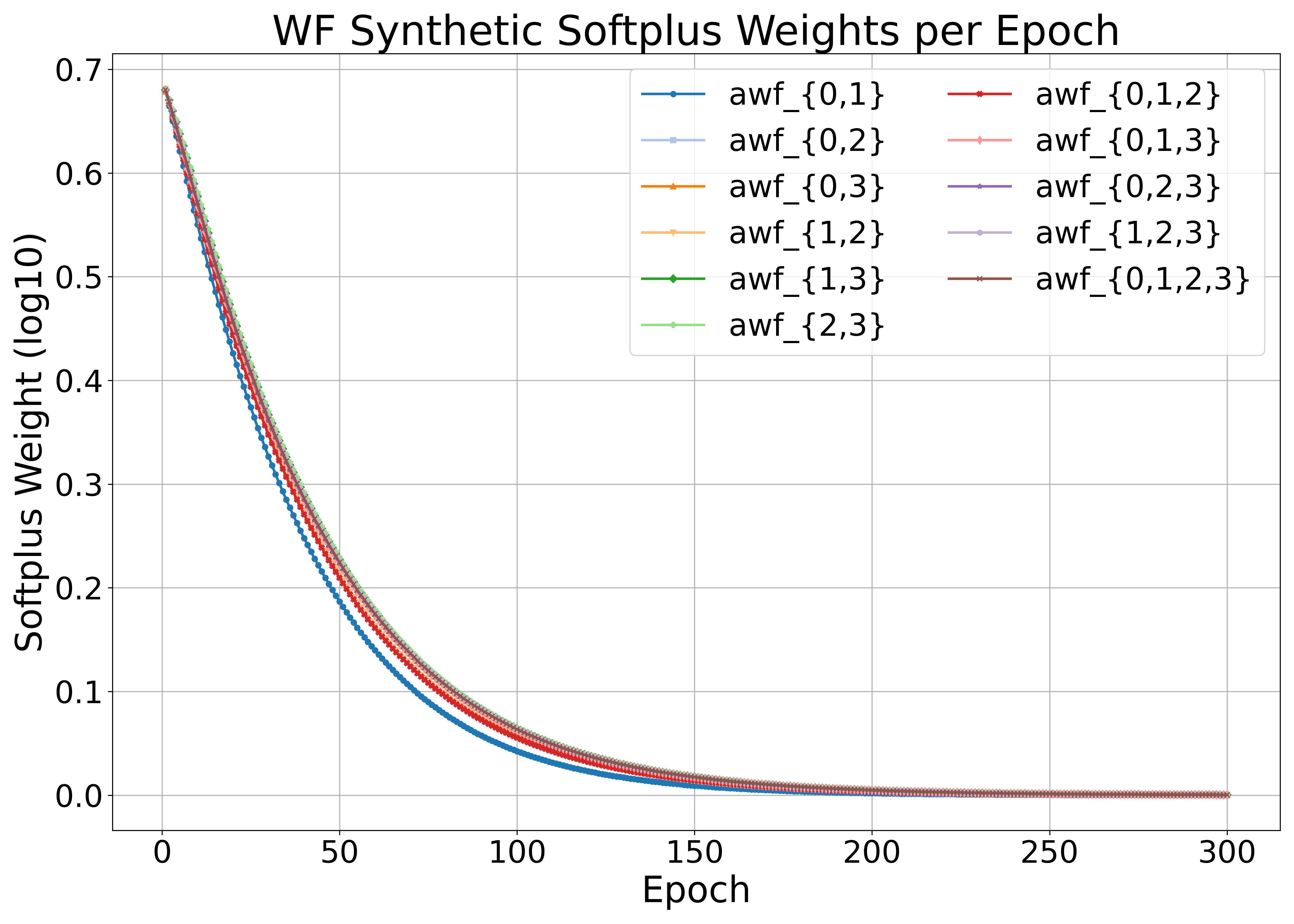}
    \caption{Linear scale}
  \end{subfigure}\hfill
  \begin{subfigure}[t]{0.49\linewidth}
    \centering
    \includegraphics[width=\linewidth]{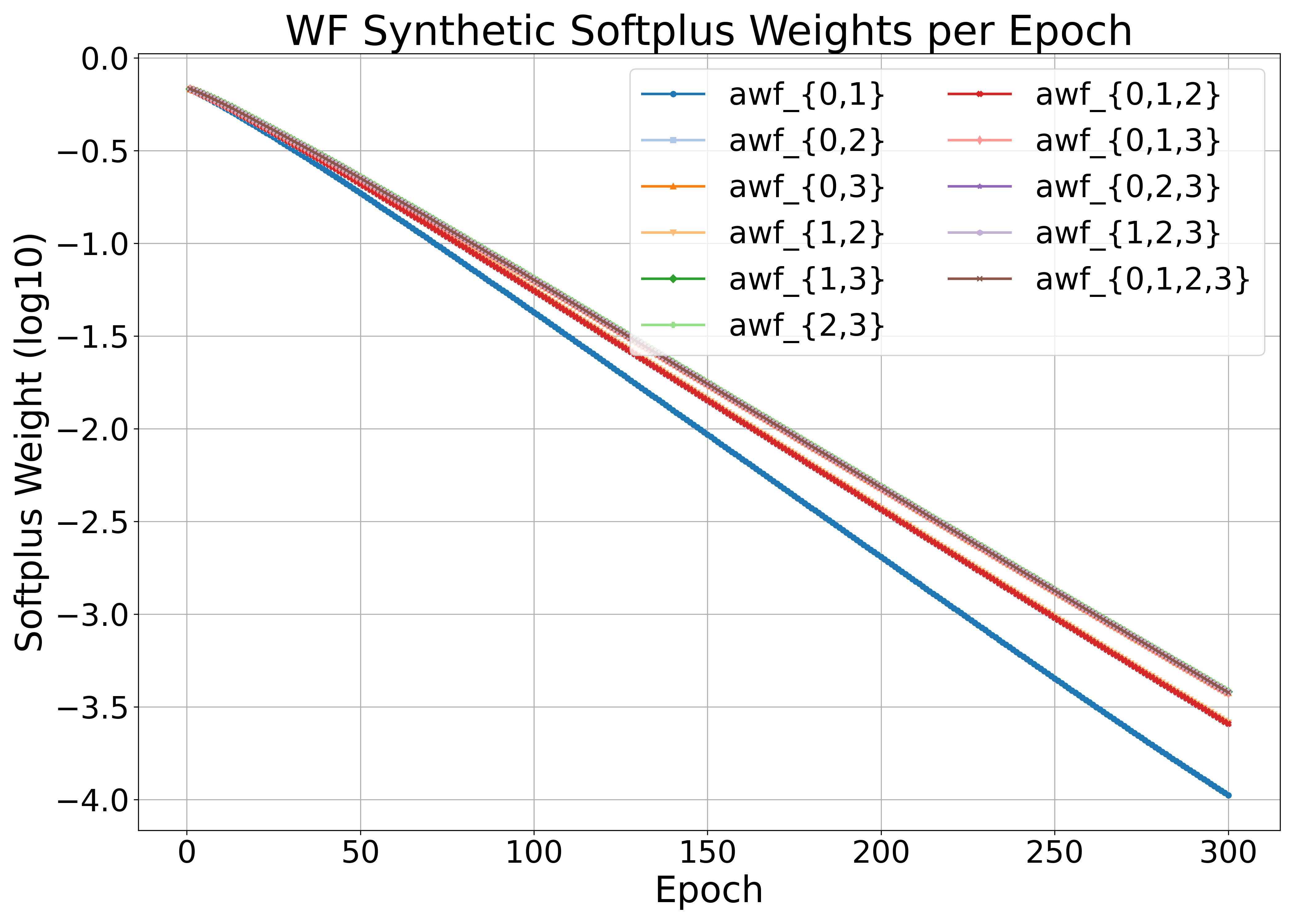}
    \caption{Log$_{10}$ scale}
  \end{subfigure}
  %
  %\vspace{-0.2cm}
  \caption{Per-subset softplus weights for the \emph{attention-weighted fusion (AWF)} operation during training. Curves correspond to every AWF mixture of decoder logits. Both the (a) linear- and (b) log–scaled plots reveal how the NAS-inspired optimization reallocates loss weight across epochs.}
  \label{fig:awf_weight_curves}
  %\vspace{-0.7cm}
\end{figure}

\textbf{Addition (Add).} Figure \ref{fig:add_weight_curves} shows that all additive subsets decay smoothly, with the full 4-logit sum ($add_{0,1,2,3}$) holding the largest weight throughout training. This indicates that the optimizer values a coarse, resolution-agnostic blending of logits for global consistency even late into training.

\textbf{Multiplication (Mult).} Figure \ref{fig:mul_weight_curves} demonstrates that multiplicative subsets start at the same magnitude as Add, leaving the 3- and 4-branch products dominant after $~$150 epochs. This pattern suggests that the optimizer relies on multiplicative interactions primarily when multiple scales jointly agree, using them as a selective gating rather than an expansive fusion.

\textbf{Concatenation (Concat).} All Concat curves in Figure \ref{fig:concat_weight_curves} lie almost on top of each other, revealing that concatenation quickly learns to down-weight raw channel stacks. The uniform and steady decay implies that concatenation contributes mainly in the early epochs to speed convergence, with its influence reduces as finer operators take over.

\textbf{Attention-Weighted Fusion (AWF).} Figure \ref{fig:awf_weight_curves} shows that AWF subsets preserve comparatively higher weights for longer—especially the full 4-logit fusion—remaining above other operators. This persistence shows the optimizer’s strong preference for spatially adaptive weighting, thus confirming AWF’s key role in exploiting complementary decoder resolutions throughout training.

\begin{figure}[t]
  \centering
  % adjust the path to where the PNG is stored
  \includegraphics[width=\textwidth]{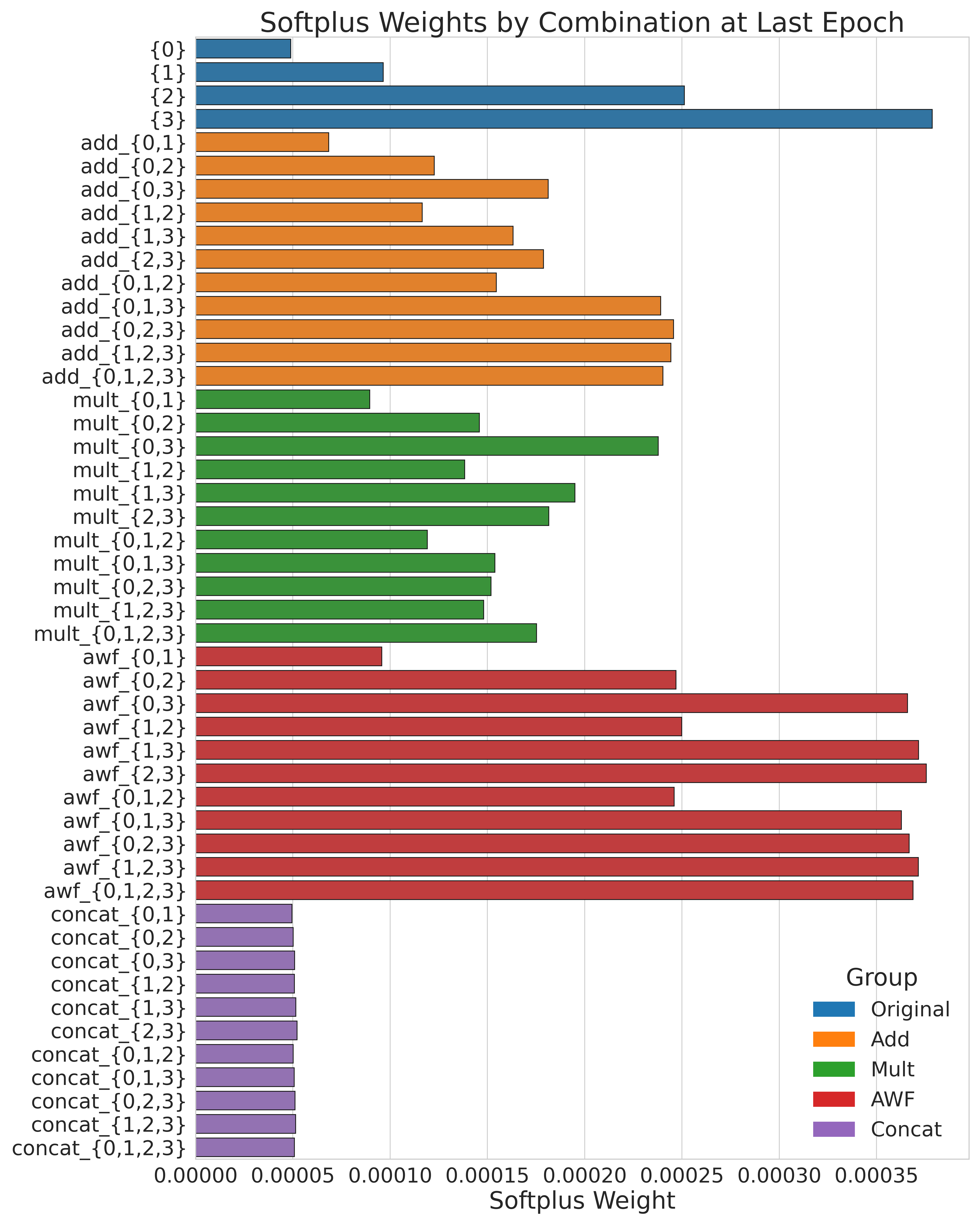}
  %\vspace{-4pt}
  \caption{Learned softplus weights of the \textbf{best} LoMix epoch. Each bar corresponds to the final softplus weight assigned to one logit combination, grouped by fusion operator (Original decoders, Addition, Multiplication, AWF, and Concatenation). Longer bars indicate that a combination is more strongly trusted by the learned loss during training.}
  \label{fig:last_epoch_weights}
    %\vspace{-0.6cm}
\end{figure}

\subsection{Learned Loss Weights Comparison of the Best Epoch}
At convergence (Figure~\ref{fig:last_epoch_weights}), the NAS-inspired optimizer concentrates most of the loss weight on attention-weighted fusion (AWF) combinations, followed by a smaller but still meaningful allocation to multiplicative and additive mixes, while concatenation paths receive almost negligible weight. Notably, among the original logits only the fine grain stages ({2}, {3}) receive higher weights, confirming that the network relies primarily on multi-scale mixtures rather than any single head. This distribution echoes our ablation study: the model learns to emphasize operators that can reconcile complementary spatial cues (AWF, Mult, Add), while down-weighting logits that add parameters without clear synergy (Concat), thereby producing the highest DICE performance without manual tuning.

%Line/heatmap plots of learned softplus weights across epochs

%Show convergence patterns and shifts in contribution

%Discuss if original or synthesized weights dominate, and how fusion types are prioritized

%\subsection{Qualitative Mutated Logits Visualization}

%Show comparisons: baseline vs. individual decoder vs. fused prediction

%Add confidence/entropy maps if available

\begin{table}[]
\begin{center}
%\vspace{-0.2cm}
\caption{Results of Synapse 8-organ segmentation. DICE scores (\%) are reported for individual organs. Results of UNet, AttnUNet, PolypPVT, SSFormerPVT, TransUNet, and SwinUNet are taken from \cite{rahman2024emcad}. Gallbladder (GB), Left kidney (KL), Right kidney (KR), Pancreas (PC), Spleen (SP), and Stomach (SM). $\uparrow$ ($\downarrow$) denotes the higher (lower) the better. `$-$' means missing data from the source. Results are averaged over five runs. Best results are shown in \textbf{bold}.}
%Note that LoMix adds no overhead at test time. LoMix achieves the best average DICE and lowest HD95, with particularly large gains on challenging small organs.
%\vspace{-0.1cm}
\label{tab:multi_organ_results}
\begin{adjustbox}{width=1\textwidth}
\begin{tabular}{l|rrr|rrrrrrrrr}
\toprule
\multirow{2}{*}{Methods} & \multicolumn{3}{c|}{Average}   &    \multicolumn{8}{c}{Per–organ DICE (\%)$\!\uparrow$}\\
\cmidrule(lr){5-12}     
& \multicolumn{1}{l}{DICE (\%)$\uparrow$} & \multicolumn{1}{l}{HD95$\downarrow$} & \multicolumn{1}{l|}{mIoU (\%)$\uparrow$} & \multicolumn{1}{l}{Aorta}                       & \multicolumn{1}{l}{GB}                    & \multicolumn{1}{l}{KL}                         & \multicolumn{1}{l}{KR}                         & \multicolumn{1}{l}{Liver}                       & \multicolumn{1}{l}{PC}                    & \multicolumn{1}{l}{SP}                    & \multicolumn{1}{l}{SM}                    \\
\midrule
UNet \cite{ronneberger2015u}                   & 70.11                    & 44.69                    & 59.39                                        & 84.00                                      & 56.70                                   & 72.41                                        & 62.64                                        & 86.98                                      & 48.73                                   & 81.48                                   & 67.96                                   
\\
AttnUNet \cite{oktay2018attention}                   & 71.70                    & 34.47                    & 61.38                                   & 82.61                                      & 61.94                                   & 76.07                                        & 70.42                                        & 87.54                                      & 46.70                                   & 80.67                                   & 67.66                                   
\\
UNet++  \cite{zhou2018unet++} & 80.01 & 28.08 & 69.91 & 89.15 & 70.99 & 83.37 & 79.21 & 94.00 & 61.23 & 86.38 & 75.79 \\

%R50+UNet \cite{chen2021transunet}                   & 74.68                    & 36.87                    & $-$                                     & 84.18                                      & 62.84                                   & 79.19                                        & 71.29                                        & 93.35                                      & 48.23                                   & 84.41                                   & 73.92                                   \\
%R50+AttnUNet \cite{chen2021transunet}                   & 75.57                    & 36.97                    & $-$                                    & 55.92                                      & 63.91                                   & 79.20                                        & 72.71                                        & 93.56                                      & 49.37                                   & 87.19                                   & 74.95                                   \\
DeepLabv3Plus-R50 \cite{chen2017deeplab} & 79.37 & 23.43 & 69.52 & 83.84 & 63.56 & 85.31 & 82.29 & 94.04 & 59.22 & 88.61 & 78.12 \\

SSFormer \cite{wang2022stepwise}                   & 78.01                    & 25.72                    & 67.23                                     & 82.78                                      & 63.74                                   & 80.72                                        & 78.11                                        & 93.53                                      & 61.53                                   & 87.07                                   & 76.61                                   \\
PolypPVT \cite{dong2021polyp}                       & 78.08                    & 25.61                    & 67.43                           & 82.34                                      & 66.14                                   & 81.21                                        & 73.78                                        & 94.37                                      & 59.34                                   & 88.05                                   & 79.4                                    \\
%TFCNs \cite{li2022tfcns}                         & 75.63                    & 30.63                    & 64.69                             & \textbf{88.23}                                      & 59.18                                   & 80.99                                        & 73.12                                        & 92.02                                      & 54.24                                   & 88.36                                   & 68.9                                    \\
TransUNet \cite{chen2021transunet}                     & 77.61                    & 26.90                     & 67.32                                  & 86.56                                      & 60.43                                   & 80.54                                        & 78.53                                        & 94.33                                      & 58.47                                   & 87.06                                   & 75.00                                      \\
SwinUNet \cite{cao2021swin}                      & 77.58                    & 27.32                    & 66.88                              & 81.76                                      & 65.95                                   & 82.32                                        & 79.22                                        & 93.73                                      & 53.81                                   & 88.04                                   & 75.79                                   \\
MT-UNet \cite{wang2022mixed}                      & 78.59                    & 26.59             & $-$       & 87.92                                      & 64.99                                   & 81.47                                        & 77.29                                        & 93.06                                     & 59.46                                  & 87.75                                   & 76.81                                  \\
MISSFormer \cite{huang2021missformer}                      & 81.96                    & 18.20             & $-$       & 86.99                                      &  68.65                                   &  85.21                                        & 82.00                                        & 94.41                                      & 65.67                                   & 91.92                                  & 80.81                                   \\ 
PVT-CASCADE \cite{Rahman_2023_WACV}                  & 81.06                    & 20.23                    & 70.88                                     & 83.01                                      & 70.59                                   & 82.23                                        & 80.37                                        & 94.08                                      & 64.43                                   & 90.10                                    & 83.69                                   \\
TransCASCADE \cite{Rahman_2023_WACV}                & 82.68                    & 17.34                    & 73.48                                         & 86.63                                      & 68.48                                   & 87.66                                        & 84.56                                        & 94.43                                     & 65.33                                   & 90.79                                   & 83.52                                   \\
%Cascaded MERIT \cite{rahman2023multi}                   & 84.32                    & 14.27    & 75.44                & 86.67                                      & 72.63                                   & 87.71                                       & 84.62                                        & 95.02                                      & \textbf{70.74}                                   &  91.98                                    & \textbf{85.17}                                   \\
Rolling-UNet-S \cite{liu2024rolling} & 74.84 & 40.07 & 63.89 & 85.43 & 65.45 & 77.18 & 71.28 & 92.50 & 49.15 & 85.77 & 72.00 \\
CMUNeXt-S \cite{tang2023cmunext} & 75.20 & 28.16 & 64.37 & 83.96 & 61.34 & 77.01 & 78.04 & 91.53 & 51.50 & 85.36 & 72.85 \\
EGE‑UNet \cite{ruan2023ege} & 62.28 & 51.22 & 48.83 & 70.43 & 51.35 & 68.28 & 59.32 & 86.70 & 42.25 & 67.06 & 52.84 \\

%$\pm$0.2
PVT-GCASCADE \cite{rahman2023g}                & 83.28                   &   15.83                  &  73.91                         &   86.50                                    &  71.71                                  &   87.07                                      &   83.77                                    &  95.31                                   &   66.72                                &   90.84                                & 83.58 \\
UNeXt \cite{valanarasu2022unext} & 72.60 & 30.68 & 61.30 & 80.20 & 60.82 & 76.13 & 69.96 & 91.80 & 48.04 & 83.27 & 70.64 \\

%MERIT-GCASCADE \cite{rahman2023g}                 & \textbf{84.54}                   & \textbf{10.38}                    & \textbf{75.83}                           &  \textbf{88.05}                                     &    \textbf{74.81}                                & \textbf{88.01}                                        & \textbf{84.83}                                        & \textbf{95.38}                                      & 69.73                                   & 91.92                                   & 83.63 \\
PVT-EMCAD-B0 \cite{rahman2024emcad}                 & 81.97                   &   17.39                  &  72.64                         &   87.21                                    &  66.62                                  &   87.48                                      &   83.96                                    &  94.57                                   &   62.00                                &   92.66                                & 81.22
\\
%PVT-EMCAD-B2 + DS & & 82.90 & 19.70 & 73.84 & 87.43 & 67.80 & 87.66 & 83.75 & 95.18 & 65.63 & 91.53 & 84.19 \\

PVT-EMCAD-B2 \cite{rahman2024emcad}                 & 83.63                   & 15.68                    & 74.65                           &  88.14                                     &    68.87                                & 88.08                                        & 84.10                                        & 95.26                                      & 68.51                                   & 92.17                                   & 83.92
\\
\midrule
PVT-EMCAD-B0 + LoMix (\textbf{Ours}) & 82.60 $\pm0.9$ & 16.80 & 73.44 & 87.41 & 68.92 & 86.67 & 83.77 & 95.41 & 62.92 & \textbf{92.70} & 83.02 \\
PVT-EMCAD-B2 + LoMix (\textbf{Ours}) & \textbf{85.07} $\pm1.3$ & \textbf{14.85} & \textbf{76.41} & \textbf{88.84} & \textbf{73.51} & \textbf{89.07} & \textbf{84.71} & \textbf{95.76} & \textbf{69.74} & 92.47 & \textbf{86.47} \\
%PVT-EMCAD-B2 + LoMix (\textbf{Ours}) & \textbf{84.91} & \textbf{14.60} & \textbf{76.21} & \textbf{88.67} & \textbf{73.51} & \textbf{88.93} & \textbf{84.68} & \textbf{95.71} & \textbf{69.57} & 91.94 & \textbf{86.24} \\
\bottomrule
\end{tabular}
\end{adjustbox}
\end{center}
%\vspace{-0.4cm}
\end{table}

\subsection{Results of Synapse 8-organ Segmentation} 

Table \ref{tab:multi_organ_results} shows that LoMix achieves clearly superior multi-organ segmentation on Synapse compared to prior CNN- and transformer-based methods. In particular, our LoMix variants attain the highest average DICE, lowest HD95, and highest mIoU of any approach. For example, the best LoMix model (85.07\%) outperforms PVT-EMCAD-B2 (83.63\%), while also exceeding TransCASCADE (82.68\%) and MISSFormer (81.96\%). The gains are especially large on small, difficult organs: e.g. LoMix significantly improves pancreas and gallbladder DICE versus previous models. Moreover, integrating LoMix into lightweight PVTv2 backbones yields consistent boosts: the PVT-EMCAD-B0 + LoMix (82.60\%) exceeds the DICE of PVT-B0-EMCAD (81.97\%), and similarly the PVT-EMCAD-B2 + LoMix (85.07\%) outperforms the PVT-EMCAD-B2 baseline (83.63\%). Crucially, LoMix is applied only during training (no extra inference cost), yet consistently pushes the SOTA across all key metrics. The reason behind the performance gain is that LoMix’s adaptive multi-scale logit fusion produces more accurate and robust abdominal organ segmentation, particularly for small, challenging structures such as the pancreas and gallbladder.

\begin{table}[t]
\begin{center}
\caption{Results of Synapse 13-organ segmentation. DICE scores (\%) are reported for individual organs. Gallbladder (GB), Left kidney (KL), Right kidney (KR), Pancreas (PC), Spleen (SP), Esophagus (Eso), Stomach (SM), Inferior Vena Cava (IVC), Portal and Splenic Veins (Veins), Left Adrenal Glands (LAG), and Right Adrenal Glands (RAG). Results are averaged over five runs. Best results per method are shown in bold. LoMix achieves the best average DICE and lowest HD95.}
\label{tab:synapse13_results}
\begin{adjustbox}{width=1\textwidth}
\begin{tabular}{l|rrr|rrrrrrrrrrrrr}
\toprule
\multirow{2}{*}{\textbf{Methods}} & \multicolumn{3}{c|}{\textbf{Average}} & \multicolumn{13}{c}{\textbf{Per–organ DICE} (\%)$\!\uparrow$} \\
\cmidrule(lr){5-17}
& DICE$\uparrow$ & HD95$\downarrow$ & mIoU$\uparrow$ & SP & KR & KL & GB & Eso & Liver & SM & Aorta & IVC & Veins & PC & RAG & LAG \\
\midrule
Last Layer & 67.11 & 13.96 & 58.06 & 90.54 & 82.79 & 86.59 & 66.82 & 71.33 & 95.46 & 80.49 & \textbf{87.51} & 80.37 & 64.54 & 66.00 & 0.00 & 0.00 \\
Deep Sup. & 70.42 & 15.20 & 60.06 & 90.47 & 81.83 & 85.71 & 67.87 & 70.40 & 95.68 & 82.10 & 86.84 & 77.29 & 65.87 & 65.41 & 0.00 & 46.03 \\
PVT-EMCAD-B2 \cite{rahman2024emcad} & 76.21 & 15.56 & 64.64 & \textbf{91.34} & 83.40 & 86.78 & 68.99 & 72.49 & 95.35 & 84.86 & 87.42 & 79.37 & 67.74 & 66.86 & \textbf{52.58} & 53.50 \\
\midrule
\multirow{1}{*}{\textbf{PVT-EMCAD-B2 }} \\+ LoMix (\textbf{Ours}) & \textbf{76.90} & \textbf{12.42} & \textbf{65.49} & 90.24 & \textbf{84.06} & \textbf{87.54} & \textbf{71.08} & \textbf{74.40} & \textbf{95.74} & \textbf{85.68} & 87.05 & \textbf{80.65} & \textbf{68.31} & \textbf{69.63} & 51.34 & \textbf{54.07} \\

\bottomrule
\end{tabular}
\end{adjustbox}
\end{center}
\vspace{-0.3cm}
\end{table}

\begin{table}[t]
%\vspace{-0.3cm}
\begin{center}
\caption{Results of breast cancer and skin lesion segmentation. We reproduce the results of SOTA methods using their publicly available implementations with our 80:10:10 train-val-test splits. The mean DICE scores (\%) of testset over five runs are reported. \#FLOPs of all the methods are reported for $256\times256$ inputs. Best results are shown in bold.}
\begin{adjustbox}{width=0.80\textwidth}
\begin{tabular}{l|r|r|r|r}
\toprule
\multirow{1}{*}{Methods} & \multirow{1}{*}{\#Params} & \multirow{1}{*}{\#FLOPs} & \multirow{1}{*}{BUSI} &  \multirow{1}{*}{ISIC2018} \\
\midrule
UNet \cite{ronneberger2015u} & 34.53M & 65.53G & 74.04 & 86.67 \\
UNet++ \cite{zhou2018unet++} & 9.16M & 34.65G & 74.76 & 87.46 \\
AttnUNet \cite{oktay2018attention} & 34.88M & 66.64G & 74.48 & 87.05 \\
DeepLabv3+ \cite{chen2017deeplab} & 39.76M & 14.92G & 76.81 & 88.64 \\
PraNet \cite{fan2020pranet} & 32.55M & 6.93G & 75.14 & 88.46 \\
UACANet \cite{kim2021uacanet} & 69.16M & 31.51G & 76.96 & 88.72 \\
SSFormer-L \cite{wang2022stepwise} & 66.22M & 17.28G & 78.76 & 90.25 \\
PolypPVT \cite{dong2021polyp} & 25.11M & 5.30G & 79.35 & 90.36 \\
TransUNet \cite{chen2021transunet} & 105.32M & 38.52G & 78.01 & 89.04 \\
SwinUNet \cite{cao2021swin} & 27.17M  & 6.20G & 77.38 & 88.66 \\
UNeXt \cite{valanarasu2022unext} & 1.47M & 0.57G & 74.71 & 87.78 \\
CMUNeXt \cite{tang2023cmunext} & 3.15M & 7.37G & 77.34 & 87.51 \\
Rolling-UNet-S \cite{liu2024rolling} & 1.78M &  2.10G & 76.38 &  87.35 \\
%EGE-UNet \cite{ruan2023ege} &  0.054M &  0.072G & 71.34 &  86.95 \\
PVT-CASCADE-B2 \cite{Rahman_2023_WACV} & 34.12M & 7.62G & 79.21 & 90.41 \\
PVT-EMCAD-B0 \cite{rahman2024emcad} & 3.92M & 0.84G & 79.80 & 90.70 \\
PVT-EMCAD-B2 \cite{rahman2024emcad} & 26.76M & 5.60G & 80.25 & 90.96 \\
\midrule
PVT-EMCAD-B0 + LoMix (\textbf{Ours}) & 3.92M & 0.84G & 80.47$\pm1.04$ & 90.77$\pm0.63$ \\
PVT-EMCAD-B2 + LoMix (\textbf{Ours}) & 26.76M & 5.60G & \textbf{81.32$\pm$1.21} & \textbf{91.18$\pm$0.72} \\
\bottomrule
\end{tabular}
\end{adjustbox}
%\vspace{-0.4cm}
\label{tab:results_busi}
\end{center}
%\vspace{-1.2cm}
\end{table}

\begin{table}[t]
%\vspace{-0.3cm}
\begin{center}
\caption{Results of polyp segmentation. We reproduce the results of SOTA methods using their publicly available implementations with our 80:10:10 train-val-test splits. The mean DICE scores (\%) of testset over five runs are reported. Best results are shown in bold.}
\begin{adjustbox}{width=0.9\textwidth}
\begin{tabular}{l|r|r|r|r}
\toprule
\multirow{1}{*}{Methods} & \multirow{1}{*}{\#Params} & \multirow{1}{*}{Kvasir} & \multirow{1}{*}{CVC-ColonDB} &  \multirow{1}{*}{ETIS-LaribPolypDB} \\
\midrule
UNet \cite{ronneberger2015u} & 34.53M & 82.87 & 83.95 & 76.85 \\
UNet++ \cite{zhou2018unet++} & 9.16M & 83.36 & 87.88 & 77.40 \\
AttnUNet \cite{oktay2018attention} & 34.88M & 83.49 & 86.46 & 76.84 \\
DeepLabv3+ \cite{chen2017deeplab} & 39.76M & 89.06 & 91.92 & 90.73 \\
PraNet \cite{fan2020pranet} & 32.55M & 84.82 & 89.16 & 83.84 \\
UACANet \cite{kim2021uacanet} & 69.16M & 90.17 & 91.02 & 89.77 \\
SSFormer-L \cite{wang2022stepwise} & 66.22M & 91.47 & 92.11 & 90.16 \\
PolypPVT \cite{dong2021polyp} & 25.11M & 91.56 & 91.53 & 89.93 \\
TransUNet \cite{chen2021transunet} & 105.32M & 91.08 & 91.63 & 87.79 \\
SwinUNet \cite{cao2021swin} & 27.17M & 89.59 & 89.27 & 85.10 \\
UNeXt \cite{valanarasu2022unext} & 1.47M & 77.88 & 83.84 & 74.03 \\
CMUNeXt \cite{tang2023cmunext} & 3.15M & 78.41 & 83.25 & 76.12 \\
Rolling-UNet-S \cite{liu2024rolling} & 1.78M & 75.93 &  82.48 & 73.26 \\
%EGE-UNet \cite{ruan2023ege} &  0.054M & 71.52 &  76.03 & 70.68 \\
PVT-CASCADE-B2 \cite{Rahman_2023_WACV} & 34.12M & 92.05 & 91.60 & 91.03 \\
PVT-EMCAD-B0 \cite{rahman2024emcad} & 3.92M & 91.95 & 91.71 & 91.65 \\
PVT-EMCAD-B2 \cite{rahman2024emcad} & 26.76M & 92.75 & 92.31 & 92.29 \\
\midrule
PVT-EMCAD-B0 + LoMix (\textbf{Ours}) & 3.92M & 92.34$\pm0.96$ & 93.31$\pm0.86$ & 92.74$\pm$0.79 \\
PVT-EMCAD-B2 + LoMix (\textbf{Ours}) & 26.76M & \textbf{93.45$\pm$0.87} & \textbf{93.98$\pm$0.68} & \textbf{93.10$\pm$0.96} \\
\bottomrule
\end{tabular}
\end{adjustbox}
%\vspace{-0.4cm}
\label{tab:results_polyp}
\end{center}
%\vspace{-1.2cm}
\end{table}

\subsection{Results of Synapse 13-organ Segmentation}

Table \ref{tab:synapse13_results} shows that integrating LoMix into the PVT-EMCAD-B2 network provides the strongest 13-organ performance reported to date on Synapse. LoMix improves the mean DICE to 76.90\%: a gain of +9.8\% over single-head supervision and +6.5\% over uniform deep supervision, while simultaneously reducing HD95 from 15.56 to 12.42. More importantly, improvements are \textit{not} limited to one or two easy structures. Indeed, LoMix achieves the best DICE in 10 of 13 organs, including difficult small-volume classes such as the gallbladder (+4.3\%), esophagus (+3.1\%), and portal and splenic veins (+3.8\%). It also revives the previously “dead” adrenal-gland predictions, pushing DICE from 0 to 51–54\%. Larger, context-driven organs such as liver and spleen see further improvement, and aorta performance remains on par with the best prior result. These gains confirm that LoMix’s learnable, mixed-scale supervision improves both boundary-sensitive and context-dependent structures, thus delivering a uniformly stronger and more anatomically faithful segmentation without introducing any inference-time overhead.

%\subsection{Detailed Results of Evaluation with Limited Data}
%\label{ssec:limited_data_results}

%This section extends our main Section \ref{ssec:limited_data_main} by providing detailed results in Table \ref{tab:synapse_limited_data_ablation} which provides the exact numbers behind the trends visualized in Figure \ref{fig:limited_data_synapse}. Across every data-scarcity regime, LoMix not only raises the mean DICE but simultaneously lowers HD95 and improves mIoU, confirming that the accuracy gains are accompanied by better boundaries and overlap. Critically, the per-organ columns reveal that the benefits extend to \emph{all} eight organs—especially the small, low-contrast PC, SP, and SM—which validates our claim that learnable scale mixing is most valuable where context-versus-detail trade-offs are hardest.

\subsection{Results of Breast Cancer and Skin Lesion Segmentation}
Table \ref{tab:results_busi} shows the evaluation of ultrasound breast-tumour (BUSI) and dermoscopic skin lesion (ISIC2018) benchmarks and again demonstrates that LoMix can improve the DICE score of efficient networks without increasing their computations. When integrated onto the PVT-EMCAD-B2 network, LoMix improves DICE scores by +1.07\% on BUSI and +0.22\% on ISIC 2018, surpassing heavyweight designs such as DeepLabv3+ and SSFormer-L. In all cases, the gains do not require architectural changes at the test time, confirming that LoMix’s learnable mixed-scale supervision translates into tangible DICE score improvements, even for small, noise-prone medical datasets, without compromising the compactness or efficiency of the underlying model.

\subsection{Results of Polyp Segmentation}

Table \ref{tab:results_polyp} benchmarks LoMix on three challenging colon-polyp datasets against different CNN, transformer, and lightweight hybrid methods. LoMix improves mean DICE of the PVT-EMCAD-B0 network by +0.4-1.6\% on Kvasir, CVC-ColonDB, and ETIS-LaribPolypDB, thus outperforming substantially larger models such as DeepLabv3+ and SSFormer-L. Coupling LoMix with the PVT-EMCAD-B2 establishes a new SOTA on all three datasets, thus surpassing the best baseline of PVT-EMCAD-B2 by up to +1.7\% while matching its inference parameter count and runtime. Gains are achieved without modifying the network at inference, confirming that LoMix’s learnable mixed-scale supervision improves polyp delineation accuracy and data efficiency without sacrificing the compactness of the underlying architecture.

%\subsection{Qualitative Results}
%\label{ssec:qualitative_results}

%In Figure \ref{fig:qualitative}, we report the segmentation maps of breast tumors, skin lesions, polyps, and cell segmentation for representative test images. In breast tumor segmentation, UNet, UNet++, and UNeXt show greater false segmentation, while TransUNet and our MK-UNet produce near-perfect segmentation maps. Similarly, in skin lesion segmentation, UNet, ResUNet, UNet++, AttnUNet, DeepLabV3+, PraNet, SwinuNet, and UNeXt miss part of the lesion (in red rectangular box). However, UACANet, TransUNet, ACC-UNet, and our MK-UNet can segment that challenging region well. Our MK-UNet can also segment the polyp correctly, while all other methods incorrectly segment another region as a polyp. In general, our MK-UNet produces the best overlapping segmentation map in all four tasks. The reason behind this well-rounded performance by our MK-UNet with a very low computational budget is the use of multi-kernel depth-wise convolutions along with gated and local attention mechanisms.

\begin{table}[t]
\centering
\caption{Effect of input resolution on Synapse 8-organ segmentation
(\textbf{↑} higher is better, \textbf{↓} lower is better). Each row is averaged over five runs. The best results are shown in \textbf{bold}.}
\label{tab:resolution_ablation}
\begin{adjustbox}{width=\textwidth}
\begin{tabular}{l|rrr|rrrrrrrrr}
\toprule
\multirow{2}{*}{Resolution} & \multicolumn{3}{c|}{Average} & \multicolumn{8}{c}{Per–organ DICE (\%)} \\
\cmidrule(lr){2-4}\cmidrule(lr){5-12}
 & DICE ↑ & HD95 ↓ & mIoU ↑ & Aorta & GB & KL & KR & Liver & PC & SP & SM \\
\midrule
$224\times224$ & 85.07 & 14.85 & 76.41 & 88.84 & 73.51 & 89.07 & 84.71 & 95.76 & 69.74 & 92.47 & 86.47 \\
$256\times256$ & 85.45 & \textbf{12.18} & 77.05 & 89.07 & 72.90 & \textbf{89.26} & 84.88 & 95.68 & 72.65 & \textbf{92.48} & 86.71 \\
$512\times512$ & \textbf{87.25} & 14.49 & \textbf{79.52} & \textbf{91.58} & \textbf{78.05} & 88.96 & \textbf{85.73} & \textbf{96.33} & \textbf{76.96} & 92.41 & \textbf{87.98} \\
\bottomrule
\end{tabular}
\end{adjustbox}
%\vspace{-0.4cm}
\end{table}

\subsection{Effect of Input Resolution on Synapse 8-organ Segmentation}
\label{ssec:input_resolution_results}

Table \ref{tab:resolution_ablation} shows that LoMix capitalizes on every pixel it is given. When the input image is enlarged from 224$\times$224 the mean DICE rises from 85.07\% to 85.45\% while HD95 reduces by 2.7, indicating crisper boundaries at only a modest memory cost. Doubling the image again to 512$\times$512 unlocks a further leap to \textbf{87.25\%} DICE score, setting new highs on six of eight organs: gallbladder gains +4.54\%, pancreas +7.22\%, stomach +1.51\%, and even large structures such as aorta and liver surpass 91.5\% and 96.3\% DICE, respectively. The pattern confirms that LoMix’s learnable multi-scale fusion continues to integrate fine detail without over-fitting, thus scaling with resolution.

\begin{table}[t]
\centering
\caption{Comparison of different fusion operation combinations using NAS-inspired Softplus weights and PVT-EMCAD-B2 model for Synapse 8-organ segmentation. Gallbladder (GB), Left kidney (KL), Right kidney (KR), Pancreas (PC), Spleen (SP), and Stomach (SM). $\uparrow$ ($\downarrow$) indicates higher (lower) is better. LoMix achieves the highest average DICE.}
\label{tab:synapse_fusion_combinations}
\begin{adjustbox}{width=\textwidth}
\begin{tabular}{l|rrr|rrrrrrrr}
\toprule
\textbf{Operation} & DICE$\!\uparrow$ & HD95$\!\downarrow$ & mIoU$\!\uparrow$ & Aorta & GB & KL & KR & Liver & PC & SP & SM \\
\midrule
AWF & 82.91 & 19.05 & 73.88 & 88.30 & 68.04 & 87.30 & 83.13 & 95.42 & 65.85 & 91.39 & 83.84 \\
(Mult, AWF) & 83.45 & 18.84 & 74.44 & 88.00 & 68.41 & 86.50 & 81.96 & 95.95 & 69.94 & 92.43 & 84.38 \\
(Add,Mult) & 83.51 & 22.99 & 74.63 & 88.12 & 68.35 & 86.25 & 83.54 & 95.24 & 70.09 & 90.75 & 85.74 \\
Mult & 83.55 & 17.47 & 74.33 & 88.20 & 69.34 & 88.02 & 83.84 & 95.06 & 66.59 & 91.27 & 86.09 \\
Add & 83.84 & 20.15 & 74.84 & 88.04 & 73.78 & 87.77 & 83.37 & 95.35 & 68.47 & 90.18 & 83.79 \\
(Add,AWF) & 83.95 & 19.77 & 75.08 & 88.38 & 71.09 & 86.82 & 82.82 & 95.52 & 69.24 & 92.36 & 85.37 \\
(Add,Concat) & 83.96 & 15.74 & 75.07 & 88.46 & 72.06 & 87.53 & 84.12 & 95.24 & 69.18 & 90.54 & 84.52 \\
(Add,Mult,AWF) & 84.01 & 21.62 & 74.92 & 89.44 & 71.40 & 88.78 & 84.24 & 94.98 & 68.56 & 91.56 & 83.11 \\
(Add,Concat,AWF) & 84.10 & 17.42 & 75.23 & 87.65 & 71.09 & 88.24 & 84.10 & 95.88 & 68.55 & 91.45 & 85.86 \\
(Mult, Concat) & 84.23 & 16.25 & 75.49 & 88.46 & 69.21 & 89.35 & 82.82 & 95.31 & 68.81 & 92.97 & 86.89 \\
(Add,Mult,Concat) & 84.32 & 19.14 & 75.35 & 88.50 & 71.79 & 87.59 & 83.18 & 95.02 & 70.18 & 92.78 & 85.48 \\
Concat & 84.45 & 19.33 & 75.53 & 88.12 & 71.12 & 88.32 & 84.73 & 96.05 & 69.80 & 91.22 & 86.23 \\
(Mult,Concat,AWF) & 84.61 & 20.24 & 75.75 & 88.83 & 72.60 & 87.82 & 83.58 & 95.90 & 70.14 & 91.56 & 86.45 \\
(Concat,AWF) & 84.71 & 20.39 & 76.15 & 88.94 & 71.54 & 87.83 & 85.40 & 96.12 & 69.18 & 92.11 & 86.56 \\
\midrule
LoMix (\textbf{Ours}) & \textbf{85.07} & \textbf{14.85} & \textbf{76.41} & 88.84 & 73.51 & 89.07 & 84.71 & 95.76 & 69.74 & 92.47 & 86.47 \\
\bottomrule
\end{tabular}
\end{adjustbox}
\end{table}

\subsection{Detailed Results of Fusion Operation Ablation}
This section extends Section \ref{ssec:fusion_ablation} by listing the full numeric results in Table \ref{tab:synapse_fusion_combinations}. When the search space is restricted to a single operator—e.g., only the attention-weighted fusion (AWF) or only element-wise multiplication—mean DICE remains below 83.6\%. Two-operator mixtures raise performance into the mid-84\% range, and every additional operator yields a further monotonic gain because the softplus search can explore a richer set of cross-scale interactions. The best three-operator recipe (Concat + Multiply + AWF) reaches 84.71\% DICE, but the full LoMix variant, which activates \emph{all four} operators, pushes the average to \textbf{85.07\%}. Organ-wise scores show that only the complete LoMix combination provides balanced gains across the entire anatomy set, confirming our claim that operator diversity—coupled with NAS-style weight learning—is essential for fully exploiting complementary coarse-to-fine cues.

\begin{table}[t]
\centering
\caption{Effect of NAS‑inspired Softplus weight learning on Synapse 8‑organ segmentation with PVT‑EMCAD‑B2. Fixed = uniform loss weights of 1, Learned = our NAS‑inspired softplus weights. Gallbladder (GB), Left kidney (KL), Right kidney (KR), Pancreas (PC), Spleen (SP), and Stomach (SM). $\uparrow$ ($\downarrow$) denotes higher (lower) is better. Results are averaged over five runs. The best setting for each fusion is shown in bold.}
\label{tab:synapse_weight_learning}
\begin{adjustbox}{width=\textwidth}
\begin{tabular}{ll|rrr|rrrrrrrrr}
\toprule
\multirow{2}{*}{\textbf{Fusion}} & \multirow{2}{*}{\textbf{Scheme}} &
\multicolumn{3}{c|}{\textbf{Average}} &
\multicolumn{8}{c}{\textbf{Per–organ DICE} (\%)$\!\uparrow$}\\
\cmidrule(lr){6-13}
 & & DICE$\!\uparrow$ & HD95$\!\downarrow$ & mIoU$\!\uparrow$ &
 Aorta & GB & KL & KR & Liver & PC & SP & SM \\
\midrule
Deep Sup.\  & Fixed   & 82.90 & 19.70 & 73.84 & 87.43 & 67.80 & 87.66 & 83.75 & 95.18 & 65.63 & 91.53 & 84.19 \\
            & Learned & \textbf{83.17} & \textbf{15.35} & \textbf{74.23} & 87.29 & 70.13 & 87.17 & 83.74 & 95.52 & 67.66 & 90.45 & 83.40 \\
\midrule
Add         & Fixed   & 83.63 & \textbf{15.68} & 74.65 & 88.14 & 68.87 & 88.08 & 84.10 & 95.26 & 68.51 & 92.17 & 83.92 \\
            & Learned & \textbf{83.84} & 17.15 & \textbf{74.84} & 88.04 & 73.78 & 87.77 & 83.37 & 95.35 & 68.47 & 90.18 & 83.79 \\
\midrule
Multiply    & Fixed   & 82.85 & 19.16 & 73.37 & 88.61 & 67.19 & 86.62 & 82.85 & 94.93 & 68.09 & 90.27 & 84.21 \\
            & Learned & \textbf{83.55}	& \textbf{17.47}	& \textbf{74.33}	& 88.2	& 69.34	& 88.02	& 83.84	& 95.06	& 66.59	& 91.27	& 86.09 \\
\midrule
Concat      & Fixed   & 84.14 & \textbf{17.77} & 75.15 & 87.11 & 70.51 & 88.25 & 83.70 & 95.71 & 68.27 & 92.70 & 86.86 \\
            & Learned & \textbf{84.45} & 19.33 & \textbf{75.53} & 88.12 & 71.12 & 88.32 & 84.73 & 96.05 & 69.80 & 91.22 & 86.23 \\
\midrule
AWF          & Fixed   & 82.66 & 20.56 & 73.44 & 88.21 & 69.69 & 87.48 & 81.96 & 95.05 & 67.36 & 89.20 & 82.34 \\
            & Learned & \textbf{82.91} & \textbf{19.05} & \textbf{73.88} & 88.30 & 68.04 & 87.30 & 83.13 & 95.42 & 65.85 & 91.39 & 83.84 \\
\midrule
LoMix       & Fixed   & 84.62 & 18.08 & 75.72 & 88.82 & 72.35 & 87.72 & 84.10 & 95.56 & 70.46 & 91.87 & 86.08 \\
            & Learned & \textbf{85.07} & \textbf{14.85} & \textbf{76.41} & 88.84 & 73.51 & 89.07 & 84.71 & 95.76 & 69.74 & 92.47 & 86.47 \\
\bottomrule
\end{tabular}
\end{adjustbox}
%\vspace{-0.4cm}
\end{table}

\subsection{Detailed Results of NAS-Inspired Weight Learning}
%This section extends our main Section \ref{ssec:nas_weight_learning} by providing detailed results in Table \ref{tab:synapse_weight_learning}.

This section extends Section \ref{ssec:nas_weight_learning} by presenting the full numbers in Table \ref{tab:synapse_weight_learning}. Across all five supervision settings, replacing uniform loss weights with our NAS-inspired \textit{softplus} weights provides a clear net win. For plain deep supervision the learned variant improves mean DICE from 82.90\% to 83.17\% and, more importantly, reduces HD95 by > 4 point, indicating sharper boundaries. The benefit is modest, but consistent for single-operator fusions— +0.21\% DICE for Add, +0.70\% for Multiply, +0.39\% for Concat, and +0.25\% for AWF—while simultaneously lowering or matching HD95 in every case. Crucially, when the full operator pool is active, the learned weights unlock LoMix’s headroom: DICE improves from 84.62\% to a new high of \textbf{85.07\%} and HD95 falls by over 3 point, with eight-of-eight organs improving or remaining steady, thus confirming that adaptive re-weighting helps the network emphasize whichever resolutions (and fusions) are most informative for each anatomy. Because weights are pruned scalars after training, these improvements come at \emph{zero} inference cost.

\begin{table}[t]
\centering
\caption{Comparison of different supervision schemes on Synapse 8-organ segmentation across five backbones. LoMix uses NAS-inspired softplus weighting. Gallbladder (GB), Left kidney (KL), Right kidney (KR), Pancreas (PC), Spleen (SP), and Stomach (SM). Results are averaged over multiple runs. Best result per model is \textbf{bolded}.}
\label{tab:synapse_supervision_models}
\begin{adjustbox}{width=\textwidth}
\begin{tabular}{ll|rrr|rrrrrrrrr}
\toprule
\multirow{2}{*}{\textbf{Model}} & \multirow{2}{*}{\textbf{Scheme}} &
\multicolumn{3}{c|}{\textbf{Average}} &
\multicolumn{8}{c}{\textbf{Per–organ DICE} (\%)$\!\uparrow$}\\
\cmidrule(lr){6-13}
& & DICE$\!\uparrow$ & HD95$\!\downarrow$ & mIoU$\!\uparrow$ & Aorta & GB & KL & KR & Liver & PC & SP & SM \\
\midrule
PVT\_V2\_B0 & Last Layer & 75.72 & 19.95 & 66.69 & 85.78 & 66.16 & 84.98 & 80.10 & 94.08 & 30.29 & 88.70 & 75.71 \\
 & Deep Sup. & 81.38 & 21.88 & 71.66 & 86.01 & 65.18 & 87.31 & 83.03 & 94.23 & 65.14 & 90.63 & 79.52 \\
 & LoMix (\textbf{Ours}) & \textbf{82.43} & \textbf{17.32} & \textbf{73.10} & 87.42 & 68.10 & 86.84 & 83.18 & 95.35 & 62.96 & 91.84 & 83.76 \\
\midrule
PVT\_V2\_B1 & Last Layer & 79.77 & 27.30 & 69.85 & 87.15 & 65.95 & 82.84 & 79.83 & 94.26 & 60.54 & 89.77 & 77.80 \\
 & Deep Sup. & 82.23 & 25.46 & 72.60 & 85.54 & 69.00 & 85.46 & 81.02 & 95.42 & 67.90 & 90.86 & 82.68 \\
 & LoMix (\textbf{Ours}) & \textbf{83.64} & \textbf{21.05} & \textbf{74.52} & 88.20 & 69.57 & 88.22 & 83.64 & 95.04 & 69.90 & 90.58 & 84.00 \\
\midrule
PVT\_V2\_B2 & Last Layer & 80.94 & 22.89 & 71.18 & 87.14 & 68.00 & 84.87 & 81.10 & 94.63 & 63.08 & 89.82 & 78.86 \\
 & Deep Sup. & 82.90 & 19.70 & 73.84 & 87.43 & 67.80 & 87.66 & 83.75 & 95.18 & 65.63 & 91.53 & 84.19 \\
 & LoMix (\textbf{Ours}) & \textbf{85.07} & \textbf{14.85} & \textbf{76.41} & 88.84 & 73.51 & 89.07 & 84.71 & 95.76 & 69.74 & 92.47 & 86.47 \\
\midrule
ResNet18 & Last Layer	& 70.25	& 27.26	& 60.24	& 80.79	& 59.12	& 81.06	& 78.13	& 91.13	& 15.68	& 88.10	& 67.94 \\
 & Deep Sup. & 79.80 & 20.63 & 69.74 & 85.95 & 67.13 & 84.65 & 81.06 & 93.72 & 57.72 & 89.72 & 78.49 \\
 & LoMix (\textbf{Ours}) & \textbf{82.12} & \textbf{20.37} & \textbf{72.54} & 87.03 & 70.89 & 83.60 & 80.96 & 94.57 & 65.47 & 92.42 & 82.04 \\
\midrule
ResNet34 & Last Layer & 75.50 & 33.65 & 65.57 & 86.05 & 63.30 & 83.17 & 79.74 & 93.20 & 37.70 & 86.96 & 73.87 \\
 & Deep Sup. & 81.07 & 18.04 & 71.38 & 85.78 & 68.13 & 86.78 & 82.75 & 94.06 & 60.93 & 90.10 & 80.06 \\
 & LoMix (\textbf{Ours}) & \textbf{82.91} & \textbf{15.31} & \textbf{73.61} & 87.08 & 73.61 & 87.67 & 83.77 & 94.36 & 66.54 & 91.55 & 78.73 \\
\midrule
\bottomrule
\end{tabular}
\end{adjustbox}
%\vspace{-0.4cm}
\end{table}

\subsection{Detailed Results of Different Supervision Across Backbones}
%This section extends our main Section \ref{ssec:supervision_across_backbone_ablation} by providing detailed results in Table \ref{tab:synapse_supervision_models}.
This section augments Section \ref{ssec:supervision_across_backbone_ablation} with the full numbers in Table \ref{tab:synapse_supervision_models}.
Across all five backbones—including three transformer variants (PVT\_V2\_B0/B1/B2) and two purely-convolutional networks (ResNet18/34)—our LoMix shows the supervision hierarchy: \textit{Last-Layer} $<$ \textit{Deep Supervision} $<$ \textbf{LoMix}. Switching from single-head training to LoMix improves the mean DICE by +6.71\% on the lightweight PVT\_V2\_B0, by +4.13\% on the large PVT\_V2\_B2, and by a striking +11.87\% on ResNet18, while simultaneously reducing HD95 by 2.63–18.34. Organ-wise scores echo this trend: LoMix delivers the best DICE for most classes on every backbone, with the sharpest jumps on the most scale-sensitive structures (GB, PC, SM) yet still improving saturated organs such as aorta and liver towards the performance ceiling. These consistent architecture-agnostic gains confirm its value as a plug-and-play supervision method for both CNN and transformer networks.

% In your preamble:
% \usepackage{booktabs}

\begin{table}[t]
\centering
\caption{Cross‑dataset/hospital generalization of LoMiX. Using the PVT‑EMCAD‑B2 network, all models are trained for 200 epochs on the Kvasir polyp‑segmentation training set (900 images); the epoch with the best DICE (\%) on the Kvasir validation split (100 images) is saved. Then we evaluated the generalizability on three external test sets (CVC‑ClinicDB, CVC‑ColonDB, ETIS‑LaribPolypDB), without further tuning.}
\label{tab:lomix_crossdata}
\setlength{\tabcolsep}{8pt}
\renewcommand{\arraystretch}{1.05}
\footnotesize
\begin{tabular}{lccc}
\toprule
\textbf{Methods} & \textbf{CVC-ClinicDB} & \textbf{CVC-ColonDB} & \textbf{ETIS-LaribPolypDB} \\
\midrule
Last Layer        & 80.59 & 75.31 & 71.64 \\
Deep Supervision   & 81.87 & 76.16 & 75.84 \\
MUTATION           & 81.88 & 76.39 & 75.97 \\
\textbf{LoMiX (Ours)} & \textbf{83.08} & \textbf{77.70} & \textbf{77.01} \\
\bottomrule
\end{tabular}
%\vspace{-0.3cm}
\end{table}

\begin{table}%[t]
\centering
\caption{MedNeXt's \cite{roy2023mednext} performance on 3D Synapse 8-organ segmentation with Last Layer (LL), Deep Supervision (DS), MUTATION, and LoMiX. DICE scores (\%) reported for Gallbladder (GB), Left kidney (KL), Right kidney (KR), Pancreas (PC), Spleen (SP), and Stomach (SM).}
\label{tab:lomix_3dresults}
\setlength{\tabcolsep}{6pt} % column spacing
\renewcommand{\arraystretch}{1.05} % row spacing
%\footnotesize
\begin{adjustbox}{width=\textwidth}
\begin{tabular}{lccrrrrrrrr}
\toprule
\textbf{Methods} & \textbf{Avg. DICE} & \textbf{Avg. HD95} & \textbf{Aorta} & \textbf{GB} & \textbf{LK} & \textbf{RK} & \textbf{Liver} & \textbf{PC} & \textbf{SP} & \textbf{SM} \\
\midrule
MedNeXt-M\_K3 + LL        & 86.22 & 6.62 & \textbf{91.96} & 75.59 & 87.11 & 85.14 & \textbf{96.81} & \textbf{78.31} & 91.26 & 83.61 \\
MedNeXt-M\_K3 + DS        & 86.63 & 8.31 & 91.61 & 78.91 & 90.40 & 85.94 & 94.72 & 74.95 & \textbf{92.10} & \textbf{84.43} \\
MedNeXt-M\_K3 + MUTATION  & 86.84 & 6.04 & 91.72 & \textbf{79.94} & \textbf{90.97} & 86.62 & 96.58 & 76.65 & 90.55 & 81.69 \\
\textbf{MedNeXt-M\_K3 + LoMiX (Ours)} & \textbf{87.19} & \textbf{4.84} & 91.81 & 79.87 & 90.54 & \textbf{86.65} & 96.68 & 76.95 & 90.63 & 84.37 \\
\bottomrule
\end{tabular}
\end{adjustbox}
%\vspace{-0.4cm}
\end{table}

\subsection{Cross-dataset evaluation on polyp segmentation}
Clinical deployment demands robustness across scanners and sites. Our evaluation has already addressed overfitting risk by spanning multiple, heterogeneous public datasets and modalities (i.e., abdominal CT, cardiac MRI, breast ultrasound, dermoscopy, colonoscopy), each collected with different protocols and devices. Yet LoMiX improves on every dataset without manual tuning, because it actually behaves like a regularizer. In fact, LoMiX implicitly ensembles diverse multi‑scale logits only during training, thus reducing the chance of overfitting to dataset‑specific biases. 

To make this explicit, we include a cross‑dataset experiment (e.g., train on one dataset/hospital, and test on another) as shown in Table \ref{tab:lomix_crossdata}. LoMiX produces the highest DICE on every dataset which confirm its superior ability to generalize across hospitals and acquisition devices.

\subsection{3D Feasibility and Results of Different Supervision}

LoMiX is dimension‑agnostic: it operates on C‑channel class logit maps at the loss level. Extending LoMiX to 3D is straightforward: can be done simply replacing the 2D convolutions and bilinear upsampling with 3D convolutions and trilinear upsampling, everything else remains unchanged. The fusion/weighting logic remains identical, and the cost still scales with the (small) number of decoder stages (rarely exceeding five stages). When GPU memory is low, either in 2D or 3D, standard optimizations such as gradient checkpointing or caching logits on CPU further reduce compute/memory requirements without altering the algorithm, thus keeping LoMiX practical on modest hardware (< 5 GB extra GPU memory required for backpropagation to process a 96$\times$96$\times$96 volume with a four-stage network and 9 output classes).

To show the feasibility of LoMiX with 3D networks, the new preliminary results of 3D MedNeXt \cite{roy2023mednext} with LoMiX are reported in Table \ref{tab:lomix_3dresults}. Our results (in bold) demonstrate that LoMiX achieves the best average DICE and HD95 scores among all supervisions.

% In your preamble:
% \usepackage{booktabs}

\end{document}

%% file: macros.tex
\usepackage[utf8]{inputenc} % allow utf-8 input
\usepackage[T1]{fontenc}    % use 8-bit T1 fonts
\usepackage{hyperref}       % hyperlinks
\usepackage{url}            % simple URL typesetting
\usepackage{booktabs}       % professional-quality tables
\usepackage{amsfonts}       % blackboard math symbols
\usepackage{nicefrac}       % compact symbols for 1/2, etc.
\usepackage{microtype}      % microtypography
\usepackage{xcolor}         % colors
\usepackage{amsmath,bm}
\usepackage{comment}

\usepackage{graphicx}

\usepackage{multirow}
\usepackage{colortbl}
\usepackage{amssymb}
\usepackage{times}
\usepackage{epsfig}
\usepackage{boldline}
\usepackage{algorithm}            % floating wrapper
\usepackage[noend]{algpseudocode} % modern pseudocode style

\algnewcommand{\algorithmicinput}{\textbf{Input:}}
\algnewcommand{\algorithmicoutput}{\textbf{Output:}}
\algnewcommand{\Input}[1]{\item[\algorithmicinput] #1}
\algnewcommand{\Output}[1]{\item[\algorithmicoutput] #1}

\usepackage{adjustbox}
\usepackage{caption}
\usepackage{subcaption}
\usepackage{float}
\usepackage{makecell}

\input{math}

\usepackage[accsupp]{axessibility}  % Improves PDF readability for those with disabilities.

%% file: math.tex
\DeclareMathAlphabet{\mathsfit}{\encodingdefault}{\sfdefault}{m}{sl}
\SetMathAlphabet{\mathsfit}{bold}{\encodingdefault}{\sfdefault}{bx}{n}

%% file: sections/0.abstracts.tex
\begin{abstract}
%U‑shaped decoders produce logits at several scales, spatial scales, and each scale captures a different mix of coarse context and fine detail. However, existing training pipelines still handle these maps in isolation—either supervising only the final, highest‑resolution logits or deep supervision with the same hard‑coded loss weight to every scale and never explores \emph{mixed‑scale} combinations. Consequently, the model cannot exploit the complementary coarse‑to‑fine cues that emerge only when predictions are fused. 

U‑shaped networks output logits at multiple spatial scales, each capturing a different blend of coarse context and fine detail. Yet, training still treats these logits in isolation—either supervising only the final, highest‑resolution logits or applying deep supervision with identical loss weights at every scale—without exploring \emph{mixed‑scale} combinations. Consequently, the decoder output misses the complementary cues that arise only when coarse and fine predictions are fused. To address this issue, we introduce LoMix (\underline{Lo}gits \underline{Mix}ing), a Neural Architecture Search (NAS)‑inspired, differentiable plug-and-play module that \textbf{generates} new mixed‑scale outputs and \textbf{learns} how exactly each of them should guide the training process. More precisely, LoMix mixes the multi-scale decoder logits with four lightweight fusion operators: addition, multiplication, concatenation, and attention-based weighted fusion, yielding a rich set of synthetic “mutant’’ maps. Every original or mutant map is given a softplus loss weight that is co‑optimized with network parameters, mimicking a one‑step architecture search that automatically discovers the most useful scales, mixtures, and operators. Plugging LoMix into recent U-shaped architectures (i.e., PVT‑V2‑B2 backbone with EMCAD decoder) on Synapse 8‑organ dataset improves DICE by +4.2\% over single‑output supervision, +2.2\% over deep supervision, and +1.5\% over equally weighted additive fusion, all with \textbf{zero} inference overhead. When training data are scarce (e.g., one or two labeled scans, 5\% of the trainset), the advantage grows to +9.23\%, underscoring LoMix’s data efficiency. Across four benchmarks and diverse U-shaped networks, LoMiX improves DICE by up to +13.5\% over single-output supervision, confirming that learnable weighted mixed‑scale fusion generalizes broadly while remaining data efficient, fully interpretable, and overhead-free at inference. Our implementation is available at \url{https://github.com/SLDGroup/LoMix}.

\end{abstract}

%% file: sections/1.introduction.tex
%\vspace{-0.2cm}
\section{Introduction}
\label{sec:introduction}

Precise delineation of organs, tumours, and lesions underpins radiotherapy planning, volumetric assessment, and computer‑aided diagnosis. State‑of‑the‑art systems almost invariably adopt U‑shaped encoder–decoder architectures such as UNet \cite{ronneberger2015u}, UNet++ \cite{zhou2018unet++}, Attn‑UNet \cite{oktay2018attention}, TransUNet \cite{chen2021transunet}, and SwinUNet \cite{cao2021swin}. These models generate logit maps at multiple decoder resolutions: coarse maps offer global anatomical context, whereas fine maps sharpen boundaries and reveal small pathologies. %Transformer hybrids (e.g.\ TransUNet \cite{chen2021transunet}, SwinUNet \cite{cao2021swin}) extend this recipe with long‑range self‑attention but still rely on the same multi‑scale decoder and incur considerable memory overhead.

%Yet prevailing training protocols either supervise only the final, full‑resolution logits or apply deep supervision with the same hard‑coded loss weight for every scale.

Surprisingly, the prevailing training protocols ignore most of this multi‑scale richness: they either back‑propagate loss only from the final logits (single‑output supervision) or apply deep supervision (i.e., auxiliary losses on intermediate network outputs) with an identical loss weight for every scale \cite{zhou2018unet++,fan2020pranet, Rahman_2023_WACV,rahman2025mk}. This uniform treatment presumes that each resolution is equally informative for every anatomy, an assumption contradicted by clinical practice, where minute, high‑contrast structures (e.g.\ pancreas, gall‑bladder) lean heavily on fine‑scale details, whereas large homogeneous organs (e.g.\ liver) profit chiefly from coarse context. The resulting mismatch leaves scale‑specific clues under‑exploited, and the deficiency grows when labels are scarce.

Deep supervision is also \emph{isolationist} in nature as it overlooks the synergy that can emerge when coarse- and fine-grain logits are \emph{combined}. Early “mutation’’ methods attempt to bridge this gap by summing up the decoder logits with equal weights, with MERIT \cite{rahman2023multi} being such a prominent example. However, such static fusion fixes the operator, enforces equal contributions from all scales, and demands manual retuning whenever imaging protocols, organ sizes, or data volumes change. General loss‑balancing schemes developed for multi‑task learning, such as uncertainty weighting \cite{kendall2018multi} or GradNorm \cite{chen2018gradnorm}, do not capture the structured correlations inside a single‑task, multi‑scale decoder, and thus leave much of this information content untapped.

%with e\underline{x}plicit weights
To address these limitations, we introduce \textbf{LoMix} (\underline{Lo}gits \underline{Mix}ing) to convert passive deep supervision into an \emph{active, learnable ensemble} of mixed‑scale predictions. During training every pair of decoder stages is fused by four lightweight, differentiable operators: pixel‑wise addition, multiplication, concatenation followed by a $1 \times 1$ convolution, and attention-based weighted fusion, resulting in a rich family of “mutant’’ logits that explicitly blend coarse context with fine detail. Each original or mutant map is modulated by a soft‑plus weight optimized jointly with the network, so the model performs a Neural Architecture Search (NAS) style selection of the most informative scales and fusion modes \emph{within} the main optimization loop; that is, no extra optimizer, no validation‑set grid search is needed. The added parameters are negligible and used only during training, leaving the FLOPs, latency, and memory footprint at test-time unchanged, while providing a data‑driven fusion that adapts to organ size, image contrast, and label scarcity.

When integrated into a recent U-shaped network, PVT‑V2‑B2 backbone \cite{wang2022pvt} with EMCAD decoder \cite{rahman2024emcad}, LoMix improves the mean DICE score on Synapse 8‑organ segmentation by \textbf{+4.2\%} over single‑output supervision, \textbf{+2.2\%} over uniform deep supervision, and \textbf{+1.5\%} over equal‑weight additive fusion; of note, the gains are even larger on the harder Synapse 13‑organ segmentation task. Consistent improvements on ACDC cardiac MRI and BUSI breast‑tumour ultrasound across both CNN and transformer backbones confirm robustness of LoMix. Even with only 5\% of training scans available, LoMix still delivers a \textbf{+9.23\%} DICE improvement, underscoring the data efficiency.

In summary, LoMix (i) is the \textbf{first framework} that jointly optimizes \emph{which} decoder scales to mix and \emph{how} to mix them for a single‑task; (ii) \textbf{substitutes manual loss weighting} with an automatic, interpretable, NAS‑inspired weighting mechanism; (iii) introduces \textbf{zero inference overhead}; and (iv) \textbf{consistently improves performance} across datasets, backbones, and annotation budgets. By allowing networks to learn their own multi‑scale fusion strategy, LoMix offers a principled and practical advancement that puts forth a strong baseline for data‑efficient medical image segmentation.

The remainder of this paper is structured as follows: Section \ref{sec:related_work} discusses related literature, Section \ref{sec:method} details the LoMix framework, Section \ref{sec:experiments} explains experimental evaluations, Section \ref{sec:ablation_study} presents several critical ablation studies, and Section \ref{sec:conclusion} concludes our findings and suggests future research directions.

%% file: sections/2.related_works.tex
%\vspace{-0.4cm}
\section{Related Work}
\label{sec:related_work}

\textbf{Medical Segmentation Architectures:} U-shaped encoder–decoder networks with skip connections (e.g., U-Net) are the de facto architectures in medical image segmentation \cite{ronneberger2015u}; they can capture fine details via multi-scale feature maps, but are limited by the locality of convolutional operations. For example, Chen et al. note that standard U-Net struggles to model long-range dependencies, motivating hybrid designs \cite{chen2021transunet}. To address this, recent work has proposed transformer-based backbones for segmentation. TransUNet \cite{chen2021transunet} combines a CNN encoder with a Vision Transformer to learn the global context, while the decoder recovers the spatial details. Similarly, Swin-Unet \cite{cao2021swin} uses a hierarchical Swin Transformer in both encoder and decoder, demonstrating that pure-transformer U-shaped models outperform purely convolutional ones on multi-organ tasks. 

Other variants adopt the Pyramid Vision Transformer (PVT) \cite{wang2022pvt} as the encoder. For instance, CASCADE \cite{Rahman_2023_WACV} and G-CASCADE \cite{rahman2023g} use PVT encoders with novel attention- or graph-based decoders to progressively refine multi-scale features. Polyp-PVT \cite{dong2021polyp} and SSFormer \cite{wang2022stepwise} also leverage PVT backbones for polyp segmentation, incorporating hand-crafted fusion modules (e.g., cascaded fusion, camouflage, and locality decoders) to combine features across scales. 

More recently, EMCAD \cite{rahman2024emcad} introduces an efficient multi-scale convolutional attention decoder that uses depth-wise convolutions and gated attention to fuse multi-resolution features. While these transformer-based models generate rich multi-scale outputs, they typically use fixed fusion or skip-connection schemes and do not learn explicit weights for combining feature maps.

\textbf{Deep Supervision and Static Multi-Scale Fusion:} 
Deep supervision has been widely adopted to improve training of segmentation models by attaching auxiliary loss functions to intermediate layers. For example, UNet++ \cite{zhou2018unet++} employs nested skip pathways: intermediate decoder outputs are each supervised by ground truth to encourage multi-scale consistency. In practice, however, these supervision losses are typically combined with fixed rules (e.g., simple averaging, summation, etc.). 

Likewise, multi-scale fusion in many models is static. Common approaches concatenate or sum feature maps from different depths without learning the fusion weights. For example, Polyp-PVT’s \cite{dong2021polyp} design includes fixed modules to merge encoder features across levels, but these components have pre-defined roles and uniform weighting. Such static fusion schemes (even when effective) do not adaptively learn which scales or channels to emphasize, leaving the relative contributions of multi-scale features unchanged during training.

\textbf{Mutation-Based Training and Its Limitations:} Some recent methods aim to exploit multi-scale logits through loss-level ensembling. MERIT \cite{rahman2023multi} is such a notable example: it aggregates original multi-stage logits through an additive MUTATION mixing strategy. This loss aggregation ensembles logits from different scales for final training. However, the MUTATION \textit{mixes uniformly} all decoder outputs (implicitly giving equal weight of 1 to each scale) using only element-wise addition and does not include learnable parameters for fusion. Similarly, EMCAD’s \cite{rahman2024emcad} multi-scale decoding produces several parallel predictions, but these are also combined using only additive fusion when computing loss. Hence, while mutation-based training can improve robustness via implicit ensembling, it lacks a trainable mechanism to re-weight or fuse multi-scale outputs using multiple operations dynamically. %it computes multi-scale logits self-attention at multiple scales in the encoder and uses a Cascaded Attention Decoder, %In other words, the fusion operation is limited only to addition; all scales contribute equally to the loss, and the network cannot adjust these weights. 

\textbf{Adaptive Weighting and NAS-Inspired Fusion:} Learning to balance multi-scale signals has also been studied from the perspective of multi-task learning. Notably, uncertainty-weighted loss functions \cite{kendall2018multi} and GradNorm \cite{chen2018gradnorm} automatically tune loss weights across tasks. While these methods adaptively adjust the loss terms, they are designed for distinct tasks and do not directly apply to multi-scale outputs of a single task. They also do not explicitly handle feature-level fusion. Some NAS methods have explored learnable combinations of feature maps (e.g. NAS-FPN in detection \cite{ghiasi2019fpn}), but none learns operator-level fusion for multi-scale segmentation outputs. In contrast, to the best of our knowledge, LoMix is the \textit{first approach} to enable fully learnable multi-scale fusion in segmentation: it synthesizes decoder outputs combinatorially (in the style of NAS cell search \cite{nekrasov2019fast}) and learns softplus-parameterized weights to aggregate losses. This allows LoMix to adaptively emphasize relevant scales and operations during training, going beyond fixed strategies used in prior work.

%% file: sections/3.technique.tex
\section{Method}
\label{sec:method}

\begin{figure}[t]
%\vspace{-.2cm}
\includegraphics[width=\textwidth]{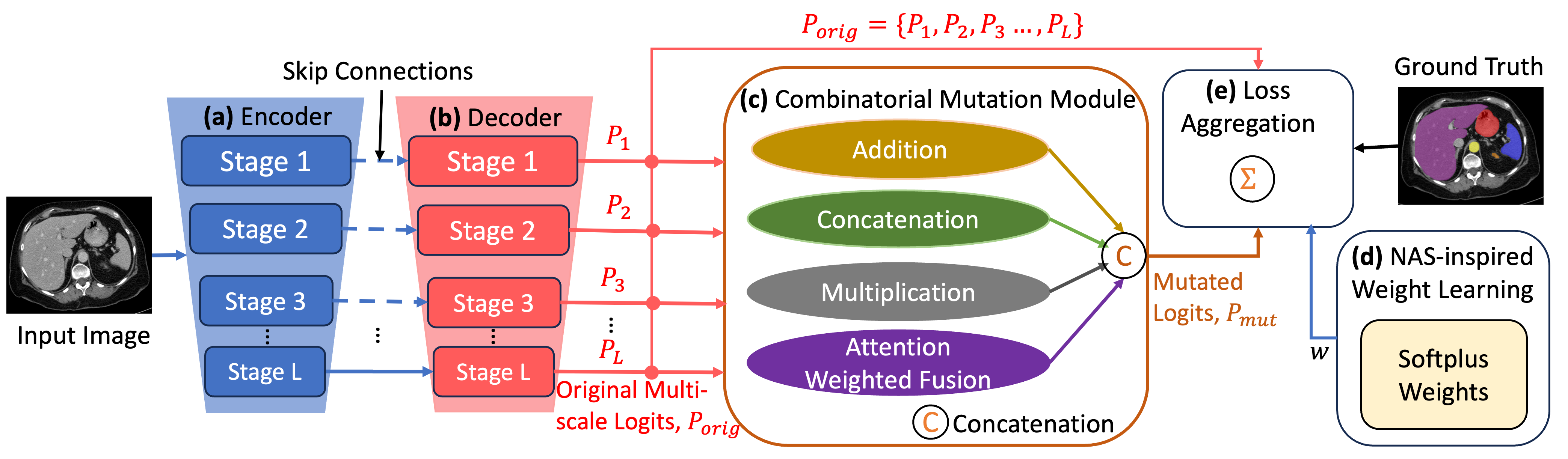}
%\vspace{-.3cm}
\caption{The proposed LoMix supervision strategy during training. \textbf{(a-b)} An input image is processed by a U–shaped network, producing original multi‑scale logits $P_{orig}$ from different stage of decoders. \textbf{(c)} The \emph{Combinatorial Mutation Module} synthesizes additional predictions by applying four fusion operators to every non‑trivial original prediction subset: \textbf{Addition}, \textbf{Concatenation}, \textbf{Multiplication}, and a learnable \textbf{Attention-Weighted Fusion (awf)}. \textbf{(d)} NAS‑inspired weight‑learning produces differentiable \emph{softplus}‑transformed weights $w$ for every original logit in $P_{orig}$ or mutated logit in $P_{mut}$, optimized jointly with network parameters via back‑propagation. \textbf{(e)} All original and mutated logits are supervised by a loss objective weighted by $\{w\}$.} %At inference time only a single final original logit map is evaluated, so \textbf{no additional computation or memory is introduced}, while the adaptive training scheme yields higher accuracy, robustness, and data‑efficiency.
%The segmentation predictions of different scales are synthesized through combinatorial operations, and their contributions are weighted using learnable parameters.} 
\label{fig:architecture}
%\vspace{-0.4cm}
\end{figure}

Next, we first review the background on U-shaped architectures and multi-scale outputs, then formalize the problem, and describe each component (see Figure~\ref{fig:architecture}).

%We propose LoMix , a plug-and-play module that converts passive deep supervision into a learnable ensemble of multi-scale logits. 
%As shown in Figure~\ref{fig:architecture}, our LoMix module takes as input the probability (logits) maps generated at each decoder stage of a U-shaped architecture \cite{ronneberger2015u, rahman2024emcad} and produces mixed-scale segmentation outputs. Our LoMix supervision approach is motivated by the observation that different decoder levels capture complementary information: lower-resolution layers have a larger receptive field capturing the global context, while higher-resolution layers preserve the fine spatial details. By training LoMix with segmentation labels, the network learns an optimal combination of these multi-scale cues. 

%\vspace{-0.2cm}
\subsection{Background: U-shaped Networks and Multi-scale Outputs}
U-shaped networks \cite{ronneberger2015u, rahman2024emcad,hassan2026efficient}, employ an encoder-decoder design with skip connections. The encoder path repeatedly downsamples the input to capture hierarchical features (see Figure \ref{fig:architecture}(a)), while the decoder path upsamples these features to reconstruct the segmentation map (see Figure \ref{fig:architecture}(b)). Skip connections between corresponding encoder and decoder layers ensure that fine spatial details lost during downsampling are recovered at the output. 

To further leverage multi-scale context, many networks produce outputs at multiple decoder stages. For example, deeply supervised networks generate intermediate segmentation maps ($P_1$, $P_2$, $P_3$,...,$P_L$ in Figure \ref{fig:architecture}) at different scales, thus providing richer gradients during training and allowing each decoder stage to capture structures at its respective scale. 

In medical imaging, this multi-scale strategy is especially beneficial, since anatomical structures can vary greatly in size. UNet++ \cite{zhou2018unet++} is one instantiation of this idea, introducing dense nested skip connections and side-output layers that fuse information across scales. Such multi-scale outputs encourage consistency across resolutions and improve medical image segmentation accuracy.

\subsection{Problem Definition: Logit Mixing (LoMix)}
Let \(X\) denote an input image of size \(H\times W\) and let \(Y\in\{1,\dots,C\}^{H\times W}\) be the ground truth with \(C\) classes. A U-shaped network with \(L\) decoder stages produces $L$ logit maps \(\{Z_\ell(X)\}_{\ell=1}^L\), where each \(Z_\ell\in\mathbb{R}^{C\times H_\ell\times W_\ell}\) has spatial resolution \(H_\ell\times W_\ell\). Upsampling each \(Z_\ell\) to the full resolution gives logit maps $P_\ell(X) \;=\;\sigma\bigl(Z_\ell(X)\bigr)
\;\in\;[0,1]^{C\times H\times W}$, with \(\sigma\) a softmax (for multi-class segmentation) or sigmoid (for binary segmentation), as appropriate. We denote the set of these original logits by \(\mathcal{P}_{\mathrm{orig}}\). LoMix also synthesizes a collection \(\mathcal{P}_{\mathrm{mut}}\) of “mutant” maps through fusion operators (addition, multiplication, concatenation, and attention-weighted fusion) in the Combinatorial Mutation Module (see Section \ref{ssec:cmm} and Figure \ref{fig:architecture}(c)), forming the combined set \(\mathcal{P}=\mathcal{P}_{\mathrm{orig}}\cup\mathcal{P}_{\mathrm{mut}}\).

For each map \(P_u\in\mathcal{P}\), we introduce a scalar \(\alpha_u\in\mathbb{R}\) and convert it to a positive loss weight
$w_u = \mathrm{softplus}(\alpha_u) = \ln\bigl(1 + e^{\alpha_u}\bigr) > 0$ by a NAS-inspired weight learning (see Section \ref{ssec:nwl} and Figure \ref{fig:architecture}(d)). Given a segmentation loss \(\mathcal{L}_{\mathrm{seg}}(P,Y)\) (e.g.\ Cross-entropy + DICE), we train both the network parameters \(\Theta\) and the loss-weight parameters \(\{\alpha_u\}\) by minimizing
$\min_{\Theta,\{\alpha_u\}} \sum_{P_u\in\mathcal{P}} w_u\;\mathcal{L}_{\mathrm{seg}}\bigl(P_u(X),\,Y\bigr)$ (see Section \ref{ssec:loss_aggregation} and Figure \ref{fig:architecture}(e)).
Because only the final decoder output \(P_L\) is used at the test time, LoMix adds zero inference overhead while adaptively learning which scales and fused combinations are most informative for robust segmentation. %\sum_{(X,Y)\in\mathcal{D}}

\subsection{Combinatorial Mutation Module (CMM)}
\label{ssec:cmm}
In our LoMix framework, a U‐shaped network produces \(L\) logit maps $P_1,\dots,P_L \in \mathbb{R}^{C\times H\times W}\,$,
each at a progressively finer scale (all upsampled to a common \(H\times W\) spatial size). Let \(P_i(p)\in\mathbb{R}^C\) denote the logits of $C$ classes at pixel \(p\in\{1,\dots,H\}\times\{1,\dots,W\}\) from the \(i\)-th decoder output. The Combinatorial Mutation Module (CMM) creates additional fused logits by combining subsets of these logits under four operators.  Specifically, for every non‐trivial subset \(S\subseteq\{1,\dots,L\}\) with \(\lvert S\rvert\ge2\), we define \textit{fused logit maps} \(P_S^{(\mathrm{op})}(p)\in\mathbb{R}^C\) (for \(\mathrm{op}\in\{add,mult,cat,awf\}\)) as follows:
\begin{itemize}
%\vspace{-0.2cm}
    \item \textbf{Addition (add):} We combine each subset $S$ of the original logit maps by \textit{element-wise addition} to produce fused map \(P_S^{(add)}(p)\) as in Eq. \ref{eq:addition_fusion}:
    \begin{equation}
P_S^{(add)}(p) \;=\; \sum_{\,i\in S\,}P_i(p)
\label{eq:addition_fusion}
\end{equation}
Intuitively, Addition fusion aggregates confidence from each subset of logit maps. Addition fusion will highlight regions where either decoder is confident (acting like an OR operation).
%In practice, we can average the sum, but since each $P_i$ is a probability map, their sum before the final sigmoid/softmax still reflects combined evidence.
%\vspace{-0.4cm}
\item \textbf{Multiplication (mult):} We take the \textit{element-wise (Hadamard) product} \(P_S^{(mult)}(p)\) of each subset $S$ of the original logit maps as in Eq. \ref{eq:product_fusion}:%\vspace{-0.15cm}
\begin{equation}
P_S^{(mult)}(p) \;=\; \prod_{\,i\in S\,}P_i(p)
\label{eq:product_fusion}
\end{equation}
Multiplication fusion provides high confidence only where \emph{all} logit maps agree (analogous to an AND operation). This fusion thus focuses on the intersection of the decoders' predictions, which can enhance precision by reinforcing common correct predictions and canceling out disagreements (if either logit map is uncertain, the product lowers confidence). %\vspace{-0.1cm}
\item \textbf{Concatenation (cat):} We concatenate each subset $S$ of the original logit maps \textit{channel-wise} and apply a $1\times1$ convolution to fuse them to produce fused map $P_S^{(\mathrm{cat})}(p)$ as in Eq. \ref{eq:concat_fusion}:\vspace{-0.15cm}
\begin{equation}
P_S^{(\mathrm{cat})}(p)
\;=\;
W_S\;\bigl[\,P_i(p)\bigr]_{i\in S}
\label{eq:concat_fusion}
\end{equation}
where \(\bigl[P_i(p)\bigr]_{i\in S}\in\mathbb{R}^{|S|C}\) is the channel‐wise concatenation and \(W_S\in\mathbb{R}^{C\times(|S|C)}\) is a \(1\times1\) convolution weight matrix (with output dimension equal to one logit map). This operation allows the network to learn an optimal pixel-wise linear combination of each subset of logits. The $1\times1$ convolution can be viewed as automatically weighting and combining the two inputs for each output class, potentially learning to trust one decoder more in certain regions and the other decoder elsewhere, based on data. %\vspace{-0.1cm}
\item \textbf{Attention-Weighted Fusion (awf):} We introduce an \textit{attention gating} to adaptively mix each subset $S$ of original logits. We first compute attention scores and normalize using Eq. \ref{eq:attention_awf}: %\vspace{-0.15cm}
\begin{equation}
\tilde\alpha_{S}(p)
=\,
W'_S\;\bigl[P_i(p)\bigr]_{i\in S},
\quad
\alpha_{i,S}(p)
=\;
\frac{\exp\bigl(\tilde\alpha_{i,S}(p)\bigr)}
{\sum_{j\in S}\exp\bigl(\tilde\alpha_{j,S}(p)\bigr)}
\label{eq:attention_awf}
\end{equation}
Then, we take the attention-weighted sum $P_S^{(\mathrm{wf})}(p)$ of \textit{each subset of logits} as in Eq. \ref{eq:awf_fusion}: \vspace{-0.15cm}
\begin{equation}
P_S^{(\mathrm{awf})}(p)
=\;
\sum_{\,i\in S\,}\alpha_{i,S}(p)\,P_i(p)
\label{eq:awf_fusion}
\end{equation}
Attention-Weighted Fusion can learn to favor the logit that is likely to be correct at each pixel of the image (for instance, one pixel might be better at fine details, another at coarse structure, so the attention gate can interpolate accordingly). It generalizes the addition fusion by allowing spatially varying weighting instead of a fixed linear mix at each pixel.

\end{itemize}
These four fusion operations help us produce \textit{new segmentation predictions} from each subset of original logits without adding significant computation (each is a simple pixel-wise operation or a $1\times1$ conv). They are complementary: addition and multiplication are fixed arithmetic mixes (one expansive, other selective), while concatenation and attention are learnable mixes (one globally learned weight, another dynamically learned per-pixel weight). By supervising all of them, we expose the network to a wide variety of joint-decoder behaviors. The decoder stages are incentivized to cooperate because an error from one decoder stage can be corrected by another in a fused output, thus leading to an overall more accurate ensemble of logits. In our LoMix framework, we apply all four fusion operations to \textit{every subset} of original logits. Adding all non‑empty subsets of the $L$ decoder predictions introduces only \(2^L-1-L\) extra logit maps (e.g., $11$ for $L=4$, $26$ for $L=5$); the total number of fused (mutant) logits is $4\bigl(2^L - 1 - L\bigr)$. Combined with the \(L\) original logits, the overall count is upper bounded by $L \;+\; 4\bigl(2^L - 1 - L\bigr)$, which remains tractable for typical \(L\le5\) since U‑shaped networks rarely exceed five stages as in \cite{ronneberger2015u, Rahman_2023_WACV, rahman2024emcad, huang2021missformer}.

%All original and fused logits are then gated by individual softplus weights and supervised with the common segmentation loss (e.g.\ cross‐entropy + DICE), yielding the final CMM objective (cf.\ Eq.~\ref{eq:total_loss}).
%which is also directly penalized by the loss. This mechanism drives the decoders to specialize and correct each other’s mistakes, leading to an overall more accurate ensemble of predictions. 

\subsection{NAS-Inspired Weight Learning}
\label{ssec:nwl}
To enable the network to learn the relative importance of each logit map, we associate a trainable scalar weight with every original decoder output $P_i$ and every fused (mutated) output $P_S^{(\text{op})}$. Concretely, let $\alpha_i\in\mathbb{R}$ be the raw parameter for output $P_i$ and $\alpha_S^{(\text{op})}\in\mathbb{R}$ the parameter for fused output $P_S^{(\text{op})}$. We map these parameters through the Softplus function to obtain strictly positive weights as in Eq. \ref{eq:softplus_weights}:%\vspace{-0.15cm}
\begin{equation}
w_i = \mathrm{softplus}(\alpha_i) = \ln(1+e^{\alpha_i}),
\quad
w_S^{(\text{op})} = \mathrm{softplus}(\alpha_S^{(\text{op})})
\label{eq:softplus_weights}
\end{equation}
By construction, $w_i>0$ and $w_S^{(\text{op})}>0$ for all $i,\text{op}$. These weights ${w_i,w_S^{(\text{op})}}$ serve as learnable scaling factors on the loss contributions of each corresponding logit map. All parameters $\alpha_i$ and $\alpha_S^{(\text{op})}$ are learned jointly with the network weights via backpropagation on the overall training objective. %In practice, we initialize these $w$ parameters to small values so that each $w\approx \ln2$, and then allow gradient descent to adapt them during training. 

These design choices allow the model to automatically learn how much emphasis to place on each original and fused logit during training, in a manner reminiscent of architecture weighting in NAS but applied to losses. \textbf{The learned weights are shown in Appendix \ref{ssec:weight_dynamics} of Supplementary Material.}

\subsection{Loss Aggregation}
\label{ssec:loss_aggregation}
Each output logits map is trained with a standard segmentation loss (e.g., Cross-Entropy and DICE losses). For the ground-truth mask $Y$ and the $i$-th original output $P_i$, we define its per-output loss as in Eq. \ref{eq:original_loss}. Similarly, per-output loss for each fused output $P_S^{(\text{op})}$ is defined in Eq. \ref{eq:mutated_loss}: %\vspace{-0.15cm}  
\begin{equation}
\mathcal{L}_i = \beta\mathcal{L}_{\mathrm{CE}}(P_i,Y) + \gamma\mathcal{L}_{\mathrm{DICE}}(P_i,Y)
\label{eq:original_loss}
\end{equation}%\vspace{-0.2cm}
\begin{equation}
\mathcal{L}_S^{(\text{op})} = \beta\mathcal{L}_{\mathrm{CE}}(P_S^{(\text{op})},Y) + \gamma\mathcal{L}_{\mathrm{DICE}}(P_S^{(\text{op})},Y)
\label{eq:mutated_loss}
\end{equation}
Here, $\mathcal{L}_{\mathrm{CE}}$ is the Cross-entropy loss weighted by $\beta$ and $\mathcal{L}_{\mathrm{DICE}}$ is the DICE loss weighted by $\gamma$ (here, $\beta$ + $\gamma$ = 1 and $\beta$,$\gamma>0$). The total training loss is then formed by weighting each output's loss by the corresponding learned softplus weight and summing as in Eq. \ref{eq:total_loss}:%\vspace{-0.2cm}
\begin{equation}
\mathcal{L}_{\mathrm{total}}
= \sum_{i=1}^L w_i\mathcal{L}_i +
\sum_{\text{op}} w_S^{(\text{op})}\mathcal{L}_S^{(\text{op})}
\label{eq:total_loss}
%\vspace{-0.15cm}
\end{equation}
Weighting the loss terms (rather than directly combining logits) provides several benefits:
%\vspace{-0.15cm}
\begin{itemize}
    \item First, it preserves distinct supervision for each output: each $P_i$ and $P_S^{(\text{op})}$ is individually trained and can receive gradients weighted by its own weights $w$. If a certain mutated logit $P_S^{(\text{op})}$ proves to be unhelpful or noisy, then the model can drive its $w^{(\text{op})}$ toward zero, effectively ignoring its loss contribution. Conversely, if an output is beneficial, its weight can be increased to emphasize it. This dynamic loss weighting is analogous to multi-task learning schemes where uncertainty or task relevance modulates loss terms. 
    \item Second, it avoids the pitfalls of weighting logits directly: mixing logits into a single prediction would blur their individual contributions and could hinder training of underperforming branches. Instead, our weighted loss formulation (Eq. \ref{eq:total_loss}) allows the network to automatically focus on useful logit outputs while minimizing the impact of less informative ones, thus leading to more effective training of the ensemble of original and mutated logits.
\end{itemize}

%% file: sections/4.experiments.tex
\begin{table}[t]
\begin{center}
%\vspace{-0.2cm}
\caption{Synapse 8-organ segmentation with Last Layer (LL), Deep Supervision (DS) \cite{zhou2018unet++}, MUTATION \cite{rahman2023multi}, and our LoMiX. DICE scores (\%) are reported for Gallbladder (GB), Left kidney (KL), Right kidney (KR), Pancreas (PC), Spleen (SP), and Stomach (SM). $\uparrow$ denotes the higher the better and $\downarrow$ denotes the lower the better. Results are averaged over at least three runs. Two-sided Wilcoxon signed-rank tests \cite{wilcoxon1992individual} indicate that LoMiX significantly outperforms LL and DS at $\alpha=0.05$. Best results are shown in \textbf{bold}.}
%Note that LoMix adds no overhead at test time. LoMix achieves the best average DICE and lowest HD95, with particularly large gains on challenging small organs.
%\vspace{-0.1cm}
\label{tab:new_multi_organ_results}
\begin{adjustbox}{width=1\textwidth}
\begin{tabular}{l|rrr|rrrrrrrrr}
\toprule
\multirow{2}{*}{Methods} & \multicolumn{3}{c|}{Average}   &    \multicolumn{8}{c}{Per–organ DICE (\%)$\!\uparrow$}\\
\cmidrule(lr){5-12}     
& \multicolumn{1}{l}{DICE (\%)$\uparrow$} & \multicolumn{1}{l}{HD95$\downarrow$} & \multicolumn{1}{l|}{mIoU (\%)$\uparrow$} & \multicolumn{1}{l}{Aorta}                       & \multicolumn{1}{l}{GB}                    & \multicolumn{1}{l}{KL}                         & \multicolumn{1}{l}{KR}                         & \multicolumn{1}{l}{Liver}                       & \multicolumn{1}{l}{PC}                    & \multicolumn{1}{l}{SP}                    & \multicolumn{1}{l}{SM}                    \\
\midrule
UNet \cite{ronneberger2015u} + LL                 & 70.1 & 44.7 & 59.4 & 84.0 & 56.7 & 72.4 & 62.6 & 87.0 & 48.7 & 81.5 & 67.9 \\
+ DS                      & 77.8 & 26.9 & 68.3 & 85.4 & 68.0 & 81.4 & 76.2 & 91.4 & 56.9 & 87.6 & 75.9 \\
+ MUTATION                & 81.5 & 26.4 & 71.8 & 89.4 & 70.5 & 85.4 & 80.4 & 94.1 & 66.3 & 88.3 & 77.5 \\
\textbf{+ LoMiX (Ours)}   & \textbf{83.6} & \textbf{24.3} & \textbf{74.6} & \textbf{90.4} & \textbf{75.3} & \textbf{86.3} & \textbf{82.5} & \textbf{94.3} & \textbf{67.9} & \textbf{91.8} & \textbf{80.2} \\
\midrule
AttUNet \cite{oktay2018attention} + LL      & 71.7 & 34.5 & 61.4 & 82.6 & 61.9 & 76.1 & 70.4 & 87.5 & 46.7 & 80.7 & 67.7 \\
+ DS & 77.9 & 29.9 & 68.1 & 85.4 & 67.5 & 81.4 & 77.4 & 91.0 & 57.2 & 87.1 & 76.3 \\
+ MUTATION        & 82.6 & 19.9 & 73.1 & 88.1 & 73.8 & \textbf{86.3} & 80.5 & 94.1 & \textbf{69.2} & 90.4 & 78.7 \\
\textbf{+ LoMiX (Ours)} & \textbf{83.0} & \textbf{19.4} & \textbf{74.1} & \textbf{90.0} & \textbf{75.5} & 84.4 & \textbf{81.4} & \textbf{94.5} & 67.2 & \textbf{91.3} & \textbf{79.7} \\
\midrule
TransUNet \cite{chen2021transunet} + LL            & 77.6 & 26.9 & 67.3 & 86.6 & 60.4 & 80.5 & 78.5 & 94.3 & 58.5 & 87.1 & 75.0 \\
+ DS                      & 82.7 & 17.3 & 73.5 & 86.6 & 68.5 & 87.7 & 84.6 & 94.4 & 65.3 & 90.8 & \textbf{83.5} \\
+ MUTATION                & 83.0 & 17.0 & 73.9 & \textbf{89.3} & 63.7 & 86.9 & 83.0 & \textbf{95.5} & \textbf{69.6} & \textbf{93.1} & 82.7 \\
\textbf{+ LoMiX (Ours)}   & \textbf{83.6} & \textbf{16.6} & \textbf{74.6} & 88.9 & \textbf{70.3} & \textbf{89.4} & \textbf{85.2} & 94.8 & 67.8 & 89.4 & 83.0 \\
\midrule
UNeXt \cite{valanarasu2022unext} + LL       & 70.5 & 29.2 & 60.1 & 81.9 & 30.6 & 80.8 & 75.8 & 92.3 & 48.4 & 84.1 & 70.3 \\
+ DS  & 72.6 & 30.7 & 61.3 & 80.2 & 60.8 & 76.1 & 70.0 & 91.8 & 48.0 & 83.3 & 70.6 \\
+ MUTATION          & 75.6 & 28.1 & 64.4 & 81.8 & \textbf{61.6} & \textbf{81.1} & 75.1 & 92.6 & 53.1 & 85.7 & 73.9 \\
\textbf{+ LoMiX (Ours)} & \textbf{76.8} & \textbf{22.7} & \textbf{66.2} & \textbf{83.8} & 60.9 & \textbf{81.1} & \textbf{78.3} & \textbf{92.8} & \textbf{56.2} & \textbf{86.6} & \textbf{74.5} \\
\midrule
PVT-CASCADE-B2 \cite{Rahman_2023_WACV} + LL        & 80.8 & 20.5 & 71.8 & 85.6 & 66.6 & 84.1 & 81.0 & 92.9 & 67.0 & 90.0 & 79.5 \\
+ DS & 81.1 & 20.2 & 70.9 & 83.0 & 70.6 & 82.2 & 80.4 & 94.1 & 64.4 & 90.1 & 83.7 \\
+ MUTATION          & 83.0 & 17.8 & 74.3 & \textbf{86.9} & 67.6 & 87.1 & 82.1 & 94.4 & 68.7 & 91.8 & \textbf{85.6} \\
\textbf{+ LoMiX (Ours)} & \textbf{84.3} & \textbf{16.4} & \textbf{75.4} & 86.5 & \textbf{72.8} & \textbf{87.4} & \textbf{84.6} & \textbf{95.6} & \textbf{70.1} & \textbf{92.6} & 84.8 \\
\midrule
PVT-EMCAD-B2 \cite{rahman2024emcad} + LL         & 80.9 & 22.9 & 71.2 & 87.1 & 68.0 & 84.9 & 81.1 & 94.6 & 63.1 & 89.8 & 78.9 \\
+ DS                      & 82.9 & 19.7 & 73.8 & 87.4 & 67.8 & 87.7 & 83.7 & 95.2 & 65.6 & 91.5 & 84.2 \\
+ MUTATION                & 83.6 & 15.7 & 74.7 & 88.1 & 68.9 & 88.1 & 84.1 & 95.3 & 68.5 & 92.2 & 83.9 \\
\textbf{+ LoMiX (Ours)}   & \textbf{85.1} & \textbf{14.9} & \textbf{76.4} & \textbf{88.8} & \textbf{73.5} & \textbf{89.1} & \textbf{84.7} & \textbf{95.8} & \textbf{69.7} & \textbf{92.5} & \textbf{86.5} \\
\bottomrule
\end{tabular}
\end{adjustbox}
\end{center}
%\vspace{-0.4cm}
\end{table}

%\vspace{-0.25cm}
\section{Experimental Evaluation}
\label{sec:experiments}

We evaluate LoMix on several medical image segmentation datasets. \textbf{Datasets, additional results and analyses including qualitative visualization are provided in the Supplementary Material}.

%\vspace{-0.15cm}
\subsection{Implementation details}
\label{ssec:impl_details}

Our methods are implemented and evaluated using Pytorch 1.11.0, operating on a single NVIDIA RTX A6000 GPU equipped with 48GB of RAM. We use the PVT-EMCAD-B2 as a default model \cite{rahman2024emcad} in our experiments with multi-scale kernels $[1\times1,3\times3,5\times5]$ and four stages unless otherwise mentioned. We consider all four operators (Addition, Multiplication, Concatenation, Attention-Weighted Fusion) with NAS-inspired learnable softplus weights in our LoMix supervision. Only the last-stage prediction from the decoder is used as the final segmentation output. Model optimization is achieved with AdamW \cite{loshchilov2017decoupled} optimizer with learning rate and weight decay set to $1e-4$.

%\vspace{-0.25cm}
\subsection{Comparison with SOTA Methods}

\textbf{Synapse 8-organ Segmentation:} Table \ref{tab:new_multi_organ_results} reports Synapse 8-organ results averaged over at least three runs, comparing single-output supervision (LL), Deep Supervision (DS) \cite{zhou2018unet++}, MUTATION \cite{rahman2023multi}, and LoMiX across six representative backbones spanning convolutional and transformer families (UNet \cite{ronneberger2015u}, AttUNet \cite{oktay2018attention}, TransUNet \cite{chen2021transunet}, UNeXt \cite{valanarasu2022unext}, PVT-CASCADE-B2 \cite{Rahman_2023_WACV}, and PVT-EMCAD-B2 \cite{rahman2024emcad}). LoMiX consistently achieves the highest average DICE and mIoU and the lowest HD95 for all backbones, improving DICE by +3–5\% on average over deep supervision (DS) and by up to +13.5\% over last-layer (LL) supervision, without any inference time overhead. The gains are especially pronounced on small or challenging organs such as gallbladder and pancreas, indicating that learnable mixed-scale fusion recovers complementary fine detail and coarse context that prior supervision schemes fail to exploit. These results demonstrate that LoMiX is a unified, lightweight, and plug-and-play training module that generalizes across diverse U-shaped decoders and transformer backbones while remaining fully compatible with standard inference pipelines.

\textbf{ACDC Cardiac Organ Segmentation:} Table \ref{tab:acdc_results} reports the cardiac organ segmentation on ACDC dataset, averaging over at least three runs. We can see that across a broad set of CNN and transformer models, our LoMix-enhanced models consistently deliver superior performance. Notably, PVT-EMCAD-B2 + LoMix attains a new peak average DICE of 92.51\%, surpassing its baseline PVT-EMCAD-B2 model (92.12\%) and all prior methods. It achieves the highest scores on every structure: RV 91.41\%, Myo 89.96\%, and LV 96.15\%. Even the lightweight PVT-EMCAD-B0 + LoMix improves over PVT-EMCAD-B0 (91.34\% → 91.69\%), matching or exceeding more complex cascaded and MERIT-based models. These gains confirm that LoMix’s learnable, multi-scale fusion yields more accurate and robust cardiac contours—particularly evident in the challenging myocardium region—while maintaining identical inference complexity.

\begin{table}[]
%\vspace{-0.2cm}
\begin{center}
\caption{Results of cardiac organ segmentation on ACDC dataset. DICE scores (\%) are reported for Right ventricle (RV), Myocardium (Myo), and Left ventricle (LV). $\uparrow$ ($\downarrow$) denotes the higher (lower) the better. Results are averaged over at least three runs. Best results are shown in \textbf{bold}.} 
\label{tab:acdc_results}
%\vspace{-0.1cm}
\begin{adjustbox}{width=0.75\textwidth}
\begin{tabular}{l|r|rrr}
\toprule
Methods        & Avg. DICE (\%) $\uparrow$    & \multicolumn{1}{l}{RV} $\uparrow$ & \multicolumn{1}{l}{Myo} $\uparrow$ & \multicolumn{1}{l}{LV} $\uparrow$\\
\midrule
UNet   \cite{ronneberger2015u}       & 87.55                        & 87.10                  & 80.63                   & 94.92                  \\
Attn\_UNet  \cite{oktay2018attention}  & 86.75                        & 87.58                  & 79.20                   & 93.47      \\        
%R50+UNet   \cite{chen2021transunet}       & 87.55                        & 87.10                  & 80.63                   & 94.92                  \\
%R50+AttnUNet  \cite{chen2021transunet}  & 86.75                        & 87.58                  & 79.20                   & 93.47                  \\
ViT+CUP \cite{chen2021transunet}   & 81.45                        & 81.46                  & 70.71                   & 92.18                 \\
%R50+ViT+CUP \cite{chen2021transunet} & 87.57                        & 86.07                  & 81.88                   & 94.75                  \\
TransUNet  \cite{chen2021transunet}       & 89.71                        & 86.67                  & 87.27                   & 95.18                  \\
SwinUNet \cite{cao2021swin}         & 88.07                        & 85.77                  & 84.42                   & 94.03                  \\
%MT$-$UNet \cite{wang2022mixed}         & 90.43                        & 86.64                  & 89.04                   & 95.62                  \\
MT-UNet \cite{wang2022mixed}         & 90.43                        & 86.64                  & 89.04                   & 95.62                  \\
MISSFormer \cite{huang2021missformer}         & 90.86                        & 89.55                  & 88.04                   & 94.99                  \\
PVT-CASCADE \cite{Rahman_2023_WACV}      & 91.46                        & 89.97                   & 88.9                   & 95.50                   \\
TransCASCADE \cite{Rahman_2023_WACV}    & 91.63                       & 90.25                   &  89.14                  & 95.50 \\
%MedT \cite{valanarasu2021medical} & 80.43	& 77.98 & 73.74 & 89.59 \\
Rolling\_UNet\_S \cite{liu2024rolling} & 87.59 & 85.02 & 83.59 &  94.17 \\
CMUNeXt \cite{tang2023cmunext} & 85.19 & 81.30 & 82.54 & 91.74 \\
UNeXt \cite{valanarasu2022unext} & 84.68 & 81.06 & 81.22 & 91.76 \\
Cascaded MERIT \cite{rahman2023multi}    & 91.85                       & 90.23                   &  89.53                  & 95.80 \\
PVT-GCASCADE \cite{rahman2023g} & 91.95                        & 90.31                  & 89.63                   & 95.91 \\
%MERIT-GCASCADE    & \textbf{92.23}                        & \textbf{90.64}                  & \textbf{89.96}                   & \textbf{96.08} \\
EGE-UNet \cite{ruan2023ege} & 80.68	& 76.6 & 75.21 & 90.23 \\
PVT-EMCAD-B0 \cite{rahman2024emcad}    & 91.34                        & 89.37                  & 88.99                   & 95.65 \\
PVT-EMCAD-B2 \cite{rahman2024emcad}    & 92.12                        & 90.65                  & 89.68                   & 96.02 \\
\midrule %\\
PVT-EMCAD-B0 + LoMix (\textbf{Ours}) & 91.69 $\pm0.51$ & 90.33 & 88.99 & 95.75 \\
PVT-EMCAD-B2 + LoMix (\textbf{Ours}) & \textbf{92.51} $\pm0.47$	& \textbf{91.41}	& \textbf{89.96}	& \textbf{96.15} \\
\bottomrule
\end{tabular}
\end{adjustbox}
\end{center}
%\vspace{-0.2cm}

\end{table}

\begin{table}[t]
\centering
\caption{Performance of LoMix on Synapse 8‑organ segmentation under limited data. Gallbladder (GB), Left kidney (KL), Right kidney (KR), Pancreas (PC), Spleen (SP), and Stomach (SM). \textbf{↑} higher is better, \textbf{↓} lower is better. Each row is averaged over five runs. The best entries for each limited data setting are shown in \textbf{bold}.}
\label{tab:synapse_limited_data_ablation}
\begin{adjustbox}{width=\textwidth}
\begin{tabular}{ll|rrr|rrrrrrrrr}
\toprule
\multirow{2}{*}{\textbf{Data Fraction}} & \multirow{2}{*}{\textbf{Scheme}} &
\multicolumn{3}{c|}{\textbf{Average}} &
\multicolumn{8}{c}{\textbf{Per–organ DICE} (\%)$\!\uparrow$}\\
\cmidrule(lr){6-13}
& & DICE$\!\uparrow$ & HD95$\!\downarrow$ & mIoU$\!\uparrow$ & Aorta & GB & KL & KR & Liver & PC & SP & SM \\
\midrule
40\% (7 scans) & Last Layer & 71.43 & 35.71 & 61.27 & 83.93 & 60.05 & 74.44 & 71.75 & 92.72 & 53.88 & 78.45 & 56.21 \\
 & LoMix (\textbf{Ours})     & \textbf{76.49} & \textbf{27.14} & \textbf{66.67} & \textbf{87.27} & \textbf{66.27} & \textbf{82.81} & \textbf{74.60} & \textbf{95.05} & \textbf{59.29} & \textbf{82.61} & \textbf{64.05} \\
 \midrule
20\% (4 scans) & Last Layer & 64.40 & 36.50 & 54.74 & 80.92 & 35.71 & 74.17 & 69.74 & 90.13 & 39.05 & 74.04 & 51.39 \\
 & LoMix (\textbf{Ours})     & \textbf{69.21} & \textbf{29.86} & \textbf{59.60} & \textbf{84.11} & \textbf{41.59} & \textbf{77.86} & \textbf{72.24} & \textbf{94.47} & \textbf{47.98} & \textbf{80.83} & \textbf{54.60} \\
 \midrule
10\% (2 scans) & Last Layer & 55.32 & 46.18 & 45.69 & 70.37 & 27.73 & 66.82 & 60.38 & 83.64 & 31.79 & 69.92 & 31.93 \\
 & LoMix (\textbf{Ours})     & \textbf{64.53} & \textbf{38.58} & \textbf{54.55} & \textbf{75.81} & \textbf{37.29} & \textbf{74.01} & \textbf{70.26} & \textbf{92.33} & \textbf{45.33} & \textbf{82.50} & \textbf{38.73} \\
 \midrule
5\% (1 scan)  & Last Layer & 37.22 & 72.21 & 28.83 & 49.14 & 23.54 & 41.24 & 41.17 & 73.77 & 10.13 & 46.42 & 12.36 \\
  & LoMix (\textbf{Ours})     & \textbf{46.45} & \textbf{56.94} & \textbf{37.36} & \textbf{57.58} & \textbf{26.92} & \textbf{50.30} & \textbf{49.48} & \textbf{80.40} & \textbf{16.44} & \textbf{65.72} & \textbf{24.75} \\
\bottomrule
\end{tabular}
\end{adjustbox}
%\vspace{-0.4cm}
\end{table}

\begin{figure}[t]
%\vspace{-.1cm}
\centering
\includegraphics[width=0.9\textwidth]{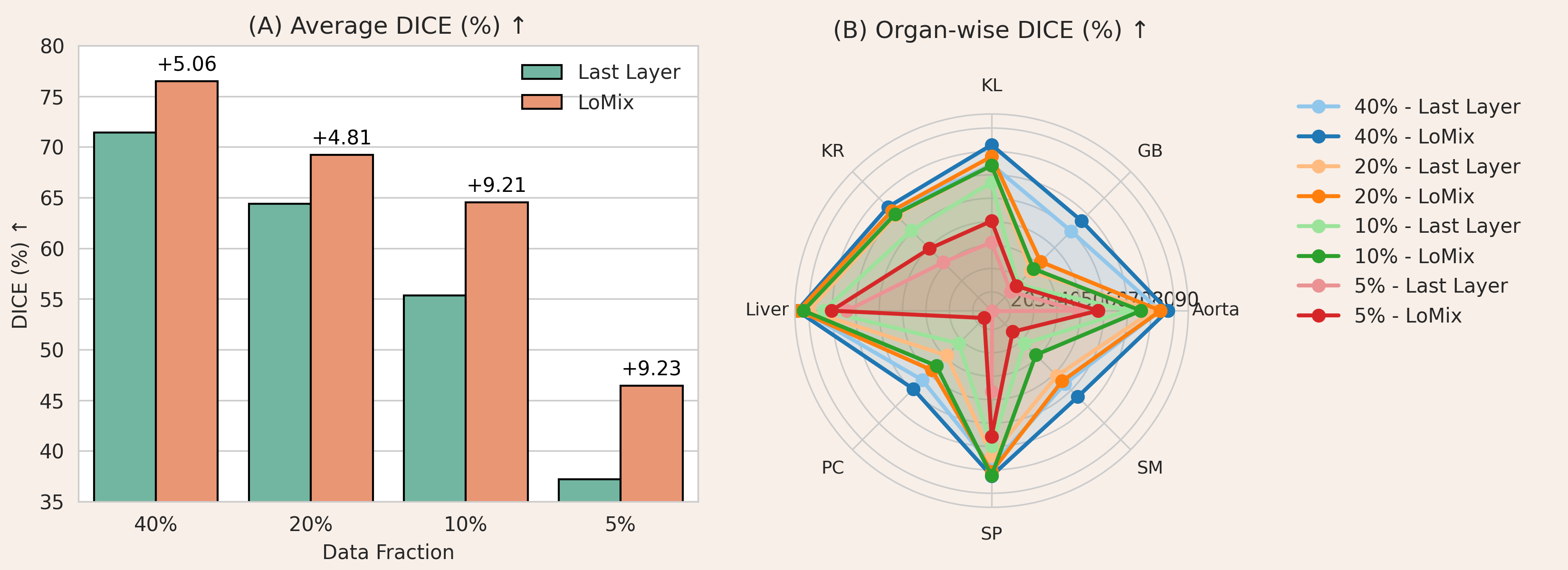}
%\vspace{-.15cm}
\caption{Limited data evaluation on Synapse 8‑organ segmentation with PVT‑EMCAD‑B2. LoMix uses NAS-inspired softplus weighting. $\uparrow$ denotes higher is better. Results are averaged over five runs. Detailed results are provided in Table \ref{tab:synapse_limited_data_ablation}.}
%The segmentation predictions of different scales are synthesized through combinatorial operations, and their contributions are weighted using learnable parameters. %and Appendix \ref{ssec:limited_data_results} in Supplementary Material} 
\label{fig:limited_data_synapse}
\vspace{-.3cm}
\end{figure}

%\vspace{-0.25cm}
\subsection{Evaluation with Limited Data}
\label{ssec:limited_data_main}

\textbf{Limited Data Setup:} We create subsets using 5\% (1 scan), 10\% (2 scans), 20\% (4 scans), and 40\% (7 scans) of the Synapse training set to evaluate performance under constrained supervision.

Figure \ref{fig:limited_data_synapse} and Table \ref{tab:synapse_limited_data_ablation} illustrate how LoMix consistently outperforms conventional single-head (“Last-Layer”) supervision as the amount of labeled data shrinks. In Table \ref{tab:synapse_limited_data_ablation} and Figure \ref{fig:limited_data_synapse}(A), with 40\% of the Synapse training set, LoMix improves the mean DICE score +5.1\%; at 20\% data the gain is +4.8\%, and when supervision drops to only 10\% and 5\% the improvement margins surge to $>$ +9\%. The radar plot on the right (Figure \ref{fig:limited_data_synapse}(B)) shows that these improvements are not confined to a single organ: LoMix raises DICE scores for every organ, with the largest boosts on small, hard-to-segment classes such as gallbladder (GB), pancreas (PC), and stomach (SM), confirming that dynamically weighting complementary scales is especially beneficial where context/details trade-offs are hardest. Hence, LoMix delivers uniform, per-organ gains and turns the U-shaped PVT-EMCAD-B2 architecture into a far more data-efficient and across-organ robust learner without any added inference cost.

%% file: sections/5.ablations.tex
\section{Ablation Study}
\label{sec:ablation_study}
This section describes three critical ablation studies. \textbf{More ablation results and analyses are provided in the Supplementary Material}.

\begin{figure}[t]
%\vspace{-.1cm}
\centering
\includegraphics[width=0.9\textwidth]{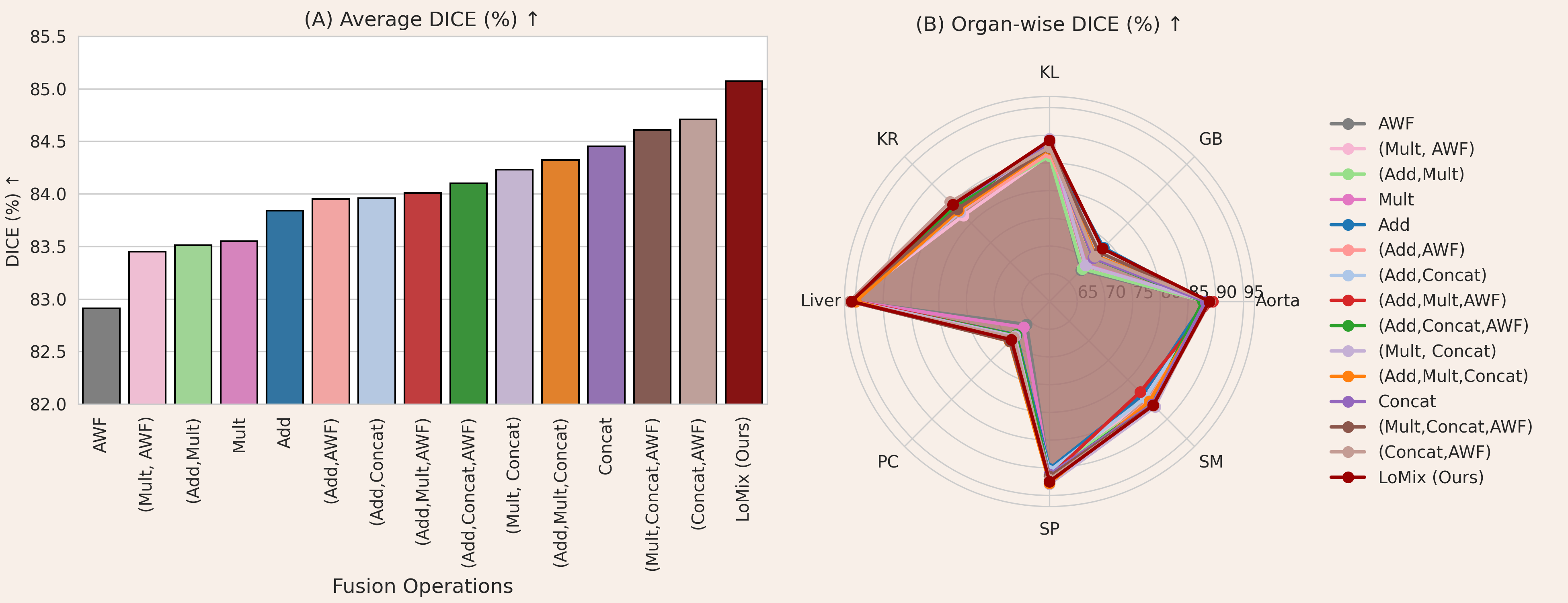}
%\vspace{-.15cm}
\caption{Comparison of different fusion operation combinations using NAS-inspired Softplus weights and PVT-EMCAD-B2 model for Synapse 8-organ segmentation. $\uparrow$ indicates higher is better.}
%The segmentation predictions of different scales are synthesized through combinatorial operations, and their contributions are weighted using learnable parameters.} 
\label{fig:operations_ablation}
\vspace{-.2cm}
\end{figure}

\begin{figure}[t]
%\vspace{-.2cm}
\centering
\includegraphics[width=0.9\textwidth]{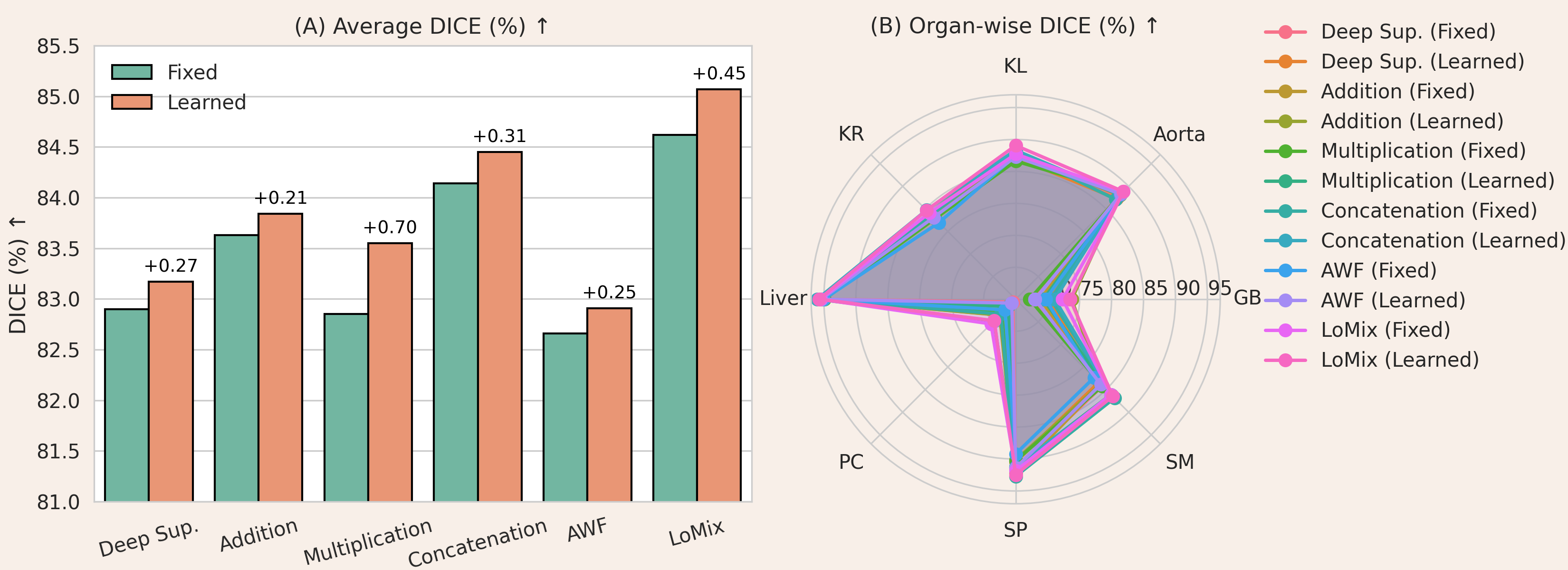}
%\vspace{-.15cm}
\caption{Effect of NAS‑inspired Softplus weight learning on Synapse 8‑organ segmentation with PVT‑EMCAD‑B2. Fixed = equal loss weights, Learned = NAS‑inspired softplus weights.}
%The segmentation predictions of different scales are synthesized through combinatorial operations, and their contributions are weighted using learnable parameters.} 
\label{fig:weight_learning}
%\vspace{-.3cm}
\end{figure}

\begin{figure}[t]
%\vspace{-.1cm}
\centering
\includegraphics[width=0.9\textwidth]{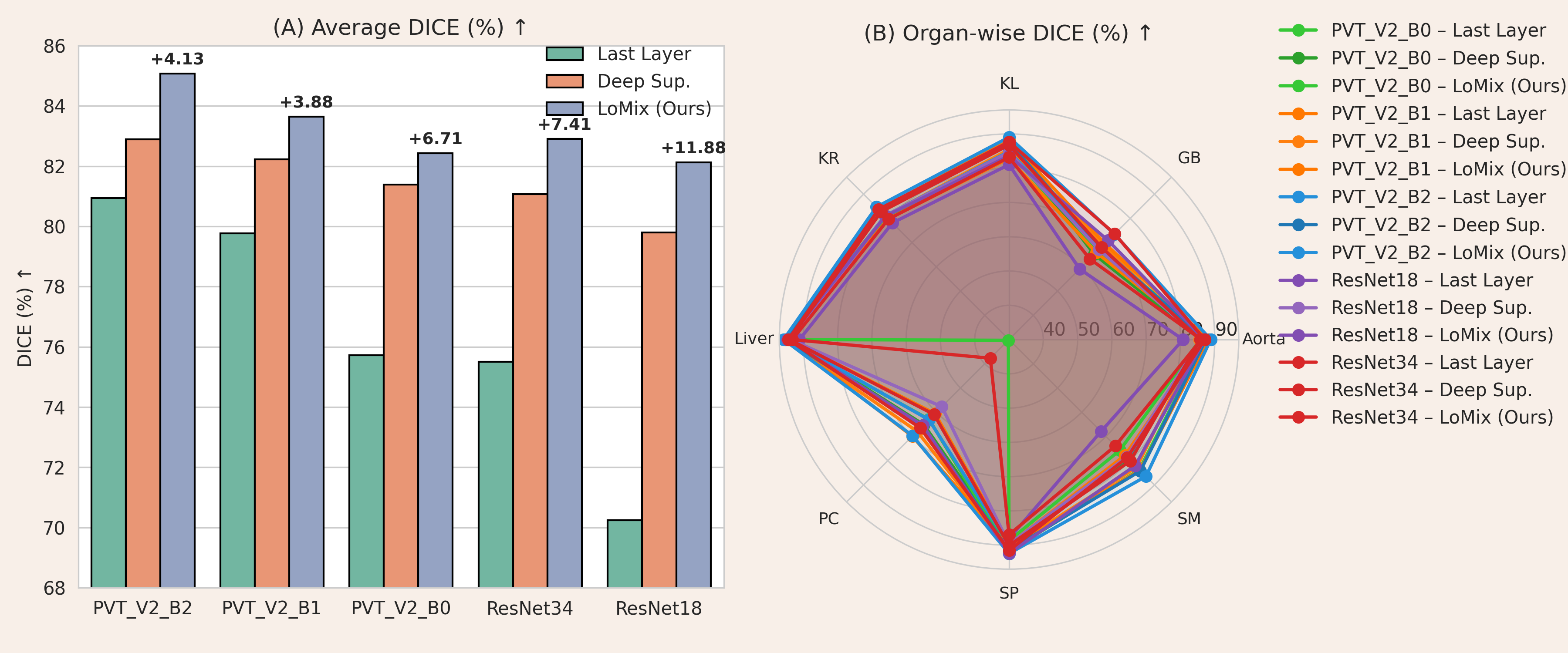}
%\vspace{-.15cm}
\caption{Comparison of different supervision schemes on Synapse 8-organ segmentation across five backbones. LoMix uses NAS-inspired softplus weighting. Sup.: Supervision.}
%The segmentation predictions of different scales are synthesized through combinatorial operations, and their contributions are weighted using learnable parameters.} 
\label{fig:supervision_across_backbone}
%\vspace{-.3cm}
\end{figure}

%\vspace{-0.2cm}
\subsection{Fusion Operation Ablation}
\label{ssec:fusion_ablation}
Figure \ref{fig:operations_ablation} shows that DICE scores increase systematically as we expand the pool of fusion operators. In Figure \ref{fig:operations_ablation}(A), starting from a single operator (AWF, leftmost bar), adding Multiply or Add provides modest gains, while injecting the Concat operation provides the sharpest jump in mean DICE. Each additional operator gives the softplus search greater freedom to discover complementary scale interactions, and the trend is strictly monotonic: the full LoMix variant that activates all four operators tops the chart at 85.07\% DICE, outperforming the best three-operator setting by ~0.46\%. The radar plot (Figure \ref{fig:operations_ablation}(B)) shows that these improvements are broader, LoMix encloses smaller polygons for every organ, with the largest margins on hard, low-contrast structures (GB, PC, SM) while still improving saturated classes such as liver and aorta toward their performance ceiling. In short, this ablation confirms that diverse operator choice, coupled with NAS-inspired weight learning, is critical: each extra operation opens a new pathway for the model to align coarse and fine cues, and the resultant mixture consistently translates into superior, organ-robust segmentation.

%\vspace{-0.2cm}
\subsection{Effect of NAS-Inspired Weight Learning}
\label{ssec:nas_weight_learning}
Figure \ref{fig:weight_learning} contrasts fixed vs. learnable NAS-inspired softplus loss weighting across six supervision types. In Figure \ref{fig:weight_learning}(A), deep supervision, single-operator fusions (addition, multiplication, concatenation, attention-weighted fusion (AWF)), and LoMix benefit from learning weights online instead of keeping them equal. The absolute gain ranges from +0.21\% DICE (Add) to a pronounced +0.7\% for Multiply, improving LoMix to 85.07\% mean DICE score without changing architecture or inference cost. The radar plot (Figure \ref{fig:weight_learning}(B)) shows that the learned variant never hurts any organ and delivers the largest jumps on the most scale-sensitive structures: gallbladder (GB) and left kidney (KL), while improving already strong classes (liver, spleen (SP)) toward the performance ceiling. Together, the results confirm that the learnable softplus weighting is a universal add-on: it tightens every supervision strategy but realizes its full potential when paired with LoMix’s rich operator pool.

%\vspace{-0.2cm}
\subsection{Effect of LoMix on Backbone Architectures}
\label{ssec:supervision_across_backbone_ablation}
Figure \ref{fig:supervision_across_backbone} demonstrates that LoMix is architecture-agnostic: whether the backbone is transformer-based (PVT-v2 \cite{wang2022pvt}) or purely convolutional (ResNet \cite{he2016deep}), replacing conventional supervision with our LoMix consistently improves performance. In Figure \ref{fig:supervision_across_backbone}(A), mean DICE score increases with supervision strength: Last Layer $<$ Deep Sup. $<$ LoMix for every network. The absolute gain delivered by LoMix over single-head supervision is generalizable: +7.41-11.88\% for ResNet variants and +3.88–6.71\% on PVT-v2 variants which shows that our LoMix supervision unlocks benefits that standard decoders leave untapped. Figure \ref{fig:supervision_across_backbone}(B) confirms that the improvements are broader: LoMix dominates other supervisions on all eight organs, with the largest margin gains again on scale-sensitive organs such as gallbladder (GB) and pancreas (PC). Crucially, these gains come at zero inference cost, thus underscoring LoMix’s practicality as a universal booster for existing segmentation backbones.

%% file: sections/6.conclusion.tex
%\vspace{-0.25cm}
\section{Conclusion}
\label{sec:conclusion}
This work introduced LoMix, a plug‑and‑play, NAS‑inspired module that unlocks the untapped value of multi‑scale decoder logits by (i) generating a rich family of mixed‑scale predictions through four differentiable fusion operators and (ii) learning, via softplus gating, how strongly each real or fused map should guide training. Extensive experiments on four medical‑image tasks demonstrate that LoMix consistently outperforms both single‑output supervision and classical deep supervision baselines while incurring \emph{no} extra inference cost. Because LoMix exposes its learned weights as explicit scalars, the fusion it learns is interpretable, transferable across backbones, and easy to fine‑tune, offering practitioners a principled yet practical path to high quality segmentation without architectural redesign. Our current implementation of LoMix is limited to 2D medical segmentation tasks. Future work will extend the approach to dense prediction tasks beyond segmentation. %explore class‑specific weighting and 
%On the widely used Synapse datasets it raises PVT-EMCAD-B2's DICE by +3.59\% on the 8‑organ segmentation and even more on the harder 13‑organ segmentation, and it delivers up to +9.23\% under extreme data scarcity, highlighting its robustness and data efficiency. 
%We introduced LogitMerge, a novel fusion and supervision framework for medical image segmentation that synthesizes and optimally weights multi-resolution decoder outputs during training. By incorporating a range of prediction fusion strategies—such as addition, multiplication, concatenation, and weighted fusion—and learning their relative importance through a differentiable, NAS-inspired loss weighting mechanism, LogitMerge addresses the limitations of static fusion and deep supervision. Our method effectively harnesses both original and synthesized predictions to enhance training dynamics and model generalization. Extensive experiments confirm that LogitMerge yields significant improvements in segmentation accuracy and robustness, even under limited data availability, while incurring no additional inference-time cost. 

%Future work will explore extending LogitMerge to multi-task segmentation scenarios and integrating uncertainty quantification for improved clinical reliability. 

\begin{ack}
This work is supported in part by the NSF grant CCF-2531882, and in part by the iMAGiNE Consortium (https://imagine.utexas.edu/).
\end{ack}

\bibliographystyle{splncs04}
\bibliography{main}